\documentclass[conference, anonymous=true]{IEEEtran}
\IEEEoverridecommandlockouts
\usepackage[nocompress,space]{cite}

\usepackage{amsmath,amssymb,amsfonts}
\usepackage{algorithmic}
\usepackage{graphicx}
\usepackage{textcomp}
\usepackage{xcolor}
\usepackage{booktabs}
\usepackage{enumitem}
\usepackage[ruled, boxed,vlined,linesnumbered]{algorithm2e}
\usepackage{verbatim}
\usepackage{subcaption}
\usepackage{hyperref}
\usepackage{balance}
\usepackage{ifthen}
\usepackage{amsthm}
\usepackage{multicol,xparse,environ}

\hypersetup{
  colorlinks   = true, 
  urlcolor     = black, 
  linkcolor    = red, 
  citecolor   = blue 
}

\let\oldnl\nl
\newcommand{\nonl}{\renewcommand{\nl}{\let\nl\oldnl}}

\def\BibTeX{{\rm B\kern-.05em{\sc i\kern-.025em b}\kern-.08em
    T\kern-.1667em\lower.7ex\hbox{E}\kern-.125emX}}

\NewEnviron{auxmulticols}[1]{%
  \ifnum#1<2\relax
    \BODY
  \else
    \begin{multicols}{#1}
      \BODY
    \end{multicols}%
  \fi
}

\begin{document}

\newboolean{isrevision}
\setboolean{isrevision}{true}

\title{Feature Inference Attack on Model Predictions in Vertical Federated Learning*\thanks{*This paper has been accepted by ICDE 2021. Yuncheng Wu is the corresponding author. }}


\author{\IEEEauthorblockN{Xinjian Luo, Yuncheng Wu, Xiaokui Xiao, Beng Chin Ooi}
	\IEEEauthorblockA{National University of Singapore\\}
	\IEEEauthorblockA{\textit{\{xinjluo, wuyc, xiaoxk, ooibc\}@comp.nus.edu.sg} \\}
}

\maketitle
\thispagestyle{plain}
\pagestyle{plain}
\begin{abstract}
Federated learning (FL) is an emerging paradigm for facilitating multiple organizations' data collaboration without revealing their private data to each other. Recently, vertical FL, where the participating organizations hold the same set of samples but with disjoint features and only one organization owns the labels, has received increased attention. This paper presents several feature inference attack methods to investigate the potential privacy leakages in the model prediction stage of vertical FL. The attack methods consider the most stringent setting that the adversary controls only the trained vertical FL model and the model predictions, relying on no background information of the attack target's data distribution. We first propose two specific attacks on the logistic regression (LR) and decision tree (DT) models, according to individual prediction output. We further design a general attack method based on multiple prediction outputs accumulated by the adversary to handle complex models, such as neural networks (NN) and random forest (RF) models.
Experimental evaluations demonstrate the effectiveness of the proposed attacks and highlight the need for designing private mechanisms to protect the prediction outputs in vertical FL.
\end{abstract}


\section{Introduction}\label{sec:introduction}

Recent years have witnessed a growing interest in exploiting data from distributed sources of multiple organizations, for designing sophisticated machine learning (ML) models and providing better customer service and acquisition. 
However, proprietary data cannot be directly shared for two reasons.
On the one hand, the usage of user’s sensitive data is compelled to abide by standard privacy regulations or laws, e.g., GDPR \cite{GDPR2016} or CCPA \cite{CCPA2018}. 
On the other hand, the data is a valuable asset to the organizations for maintaining a competitive advantage in business, which should be highly protected.

\begin{figure}[t]
    \centering
    \includegraphics[width=\columnwidth]{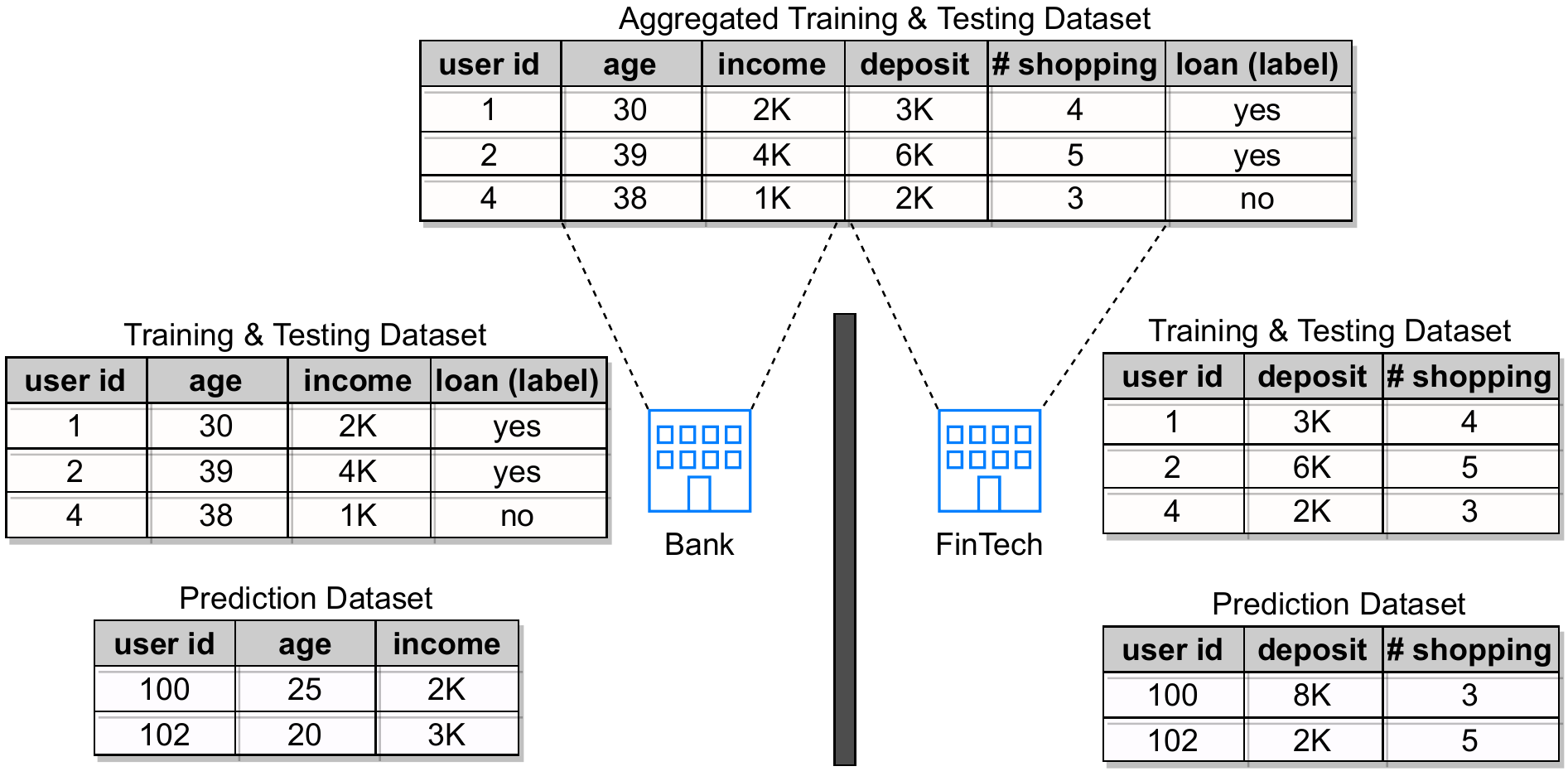}
    \caption{Example of vertical federated learning}
    \label{fig:vertical-fl-example}
    \vspace{-5mm}
\end{figure}

Federated learning (FL) \cite{WuCXCO20, Bonawitz17, McMahanMRHA17, Zheng2019, choudhury2020anonymizing, YangLCT19} is an emerging paradigm for data collaboration that enables multiple data owners (i.e., parties) to jointly build an ML model and serve new requests without revealing their private data to each other. 
%
%
FL can be categorized into different scenarios according to the data partitioning. In this paper, we consider vertical FL \cite{HuNYZ19, VaidyaCKP08, ChengCorr19, OhrimenkoSFMNVC16, WuCXCO20, HeYCGH20} where the participating parties hold the same set of samples while each party only has a disjoint subset of features.
Vertical FL has been demonstrated effective in many real-world applications and received increased attention in organizations or companies \cite{ChengCorr19, OhrimenkoSFMNVC16}. 
%
Fig.~\ref{fig:vertical-fl-example} illustrates a digital banking example. A bank aims to build an ML model to evaluate whether to approve a user's credit card application by incorporating more features from a Fintech company. 
The bank holds features of `age' and `income' while the Fintech company holds features of `deposit' and `average online shopping times'. Only the bank owns the label information in the \textit{training dataset} and \textit{testing dataset}, i.e., the ground truth that indicates whether an application should be approved. We refer to the party with the label as the \textit{active party} and the other parties as \textit{passive parties}.
To train a vertical FL model, the parties iteratively exchange certain intermediate results in a secure manner until obtaining a jointly trained model \cite{WuCXCO20}.
Finally, the trained model will be released to the parties for justifying the model's effectiveness and interpretability.

%
After obtaining the trained model, the parties utilize it to collaboratively compute model predictions for new samples in the \textit{prediction dataset}, for example, new credit card applicants in Fig.~\ref{fig:vertical-fl-example}. 
In practice, the prediction outputs will be revealed to the active party for making further decisions. 
Though cryptographic techniques, e.g., partially homomorphic encryption \cite{Paillier99, Damg01, WuWZLCL18} and secure multiparty computation \cite{Yao82b, DamgardPSZ12}, can be applied to ensure that no intermediate information during the computation is disclosed \cite{WuCXCO20}, the prediction outputs must contain information of the parties' sensitive data because they are computed upon the private data. 
Therefore, in this paper, we aim to investigate the key question: \textit{how much information about a passive party's feature values can be inferred by the active party from the model predictions in vertical FL?}


Although a number of feature inference attacks are proposed against machine learning~\cite{NasrSH2019, Carlini18, MelisSCS19, hitaj2017deep, ZhuLH19, wang2019beyond, fredrikson2014privacy, fredrikson2015model, yeom2018privacy}, none of them are applicable to our problem. 
%
%
Many existing attacks \cite{NasrSH2019, MelisSCS19, ZhuLH19, hitaj2017deep, wang2019beyond, Carlini18} aim to infer a participating party's feature values in the horizontal FL scenario \cite{Bonawitz17, McMahanMRHA17, Zheng2019, choudhury2020anonymizing}, where the parties have the same features but with different samples. 
However, these attacks greatly rely on the model gradients 
exchanged during the training process, which unintentionally memorize sensitive information of the training samples.
Once the model gradients are safely protected, e.g., using cryptographic techniques \cite{Bonawitz17, Zheng2019}, these attacks would be invalid. 
%
Also, for the inference attack in the prediction stage (i.e., considered in this paper), 
the FL model 
is not aware of the predicting samples beforehand. 
Thus, the memorization nature used in these attacks does not work. 
Another thread of feature inference attacks~\cite{fredrikson2014privacy, fredrikson2015model, yeom2018privacy} assumes that the adversary could obtain some auxiliary statistics or marginal data distribution about the unknown features. 
This assumption is relatively strong, as it is challenging in practice for the active party to collect such background information from the passive parties in vertical FL. 
In this paper, we study the privacy leakage problem in the prediction stage of vertical FL, by presenting several feature inference attacks based on model predictions. 
%
%
Our attack methods do not rely on any intermediate information during the computation of prediction outputs. For example, we assume that the parties can jointly run a secure protocol such that only the prediction outputs are released to the active party and nothing else. 
%
Besides, unlike previous work \cite{fredrikson2014privacy, fredrikson2015model, zhang2020secret}, the proposed methods need no statistics or distributions of the attack target's data.
Therefore, we consider the most stringent setting that the active party (i.e., the adversary) controls only the trained vertical FL model and the model predictions.  

%
From the experimental results, we observe that those model predictions can leak considerable information about the features held by the passive parties, especially when specific conditions are satisfied, e.g., the number of classes in the classification is large or the adversary's features and the passive parties' features are highly correlated. 
In light of these observations, we design several defense strategies and incorporate them into our \textit{Falcon}\footnote{\label{note:falcon}\url{https://www.comp.nus.edu.sg/~dbsystem/fintech-Falcon/}} (federated learning with privacy protection) system to safeguard against such potential privacy risks.

Specifically, we make the following contributions.
\begin{itemize}[topsep=2pt,itemsep=2pt,parsep=0pt,partopsep=0pt,leftmargin=15pt]
    \item We formulate the problem of feature inference attack on model predictions in vertical FL, where the active party attempts to infer the feature values of new samples belong to the passive parties. To the best of our knowledge, this is the first work that investigates this form of privacy leakage in vertical FL.
    \item We propose two specific attacks on the logistic regression (LR) and decision tree (DT) models when the adversary only has an individual prediction output. These attacks are straightforward to initiate and can achieve high accuracy when certain conditions are satisfied.
    \item We further design a general inference attack which can learn the correlations between the adversary's features and the attack target's features based on multiple predictions accumulated by the adversary. This attack can handle more complex models, such as neural networks (NN) and random forest (RF). 
    \item We implement the proposed attacks and conduct extensive evaluations on both real-world and synthetic datasets. 
    The results demonstrate the effectiveness of our attacks and highlight the need for designing defense mechanisms to mitigate the privacy risks arising from federated model predictions. 
    We also analyze and suggest several potential countermeasures against these attacks.
\end{itemize}

\section{Preliminaries}\label{sec:background}

\subsection{Machine Learning} \label{subsec:background:machine-learning}
A machine learning (ML) model is a function $f(\boldsymbol{\theta}): \mathcal{X} \rightarrow \mathcal{Y}$ represented by a set of parameters $\boldsymbol{\theta}$, where $\mathcal{X}$ denotes the input (or feature) space, and $\mathcal{Y}$ the output space \cite{MelisSCS19}. 
In this paper, we focus on supervised classification learning, which builds a model from a labeled training dataset consisting of a set of samples. 
\par 

Given a training dataset $\mathcal{D}_{\text{train}}$ with $n$ samples $\boldsymbol{x}^t (t \in \{1, \cdots, n\})$ each containing $d$ features and the corresponding output label $y^t$, to learn the parameters $\boldsymbol{\theta}$, the training algorithm optimizes a loss function, i.e., 
\begin{align}\label{eq:loss-function}
    \min_{\boldsymbol{\theta}} \frac{1}{n} \sum_{t=1}^{n} \ell (f(\boldsymbol{x}^t, \boldsymbol{\theta}), y^t) + \Omega(\boldsymbol{\theta})
\end{align}
where $\ell(f(\boldsymbol{x}^t, \boldsymbol{\theta}), y^t)$ denotes the loss of sample $\boldsymbol{x}^t$ with label $y^t$, and $\Omega(\boldsymbol{\theta})$ is the regularization term that penalizes model complexity to avoid overfitting. 

After obtaining the trained model parameters $\boldsymbol{\theta}$, we can compute the corresponding model prediction given any input sample $\boldsymbol{x}$.
In practice, the prediction output is composed of a vector of confidence scores, i.e., $\boldsymbol{v} = (v_1, \cdots, v_c)$, where each score represents the probability of $\boldsymbol{x}$ belonging to a class label, and $c$ is the number of classes.
Consequently, the classification is calculated by choosing the class label that has the highest confidence score.
\par


\vspace{1mm}\noindent
\textbf{Logistic regression (LR) model prediction.} 
In binary LR classification with two classes, the confidence score vector is $\boldsymbol{v} = (f(\boldsymbol{x}, \boldsymbol{\theta}), 1 - f(\boldsymbol{x}, \boldsymbol{\theta}))$, where
$f(\boldsymbol{x}, \boldsymbol{\theta}) = \sigma(\boldsymbol{\theta}^T \boldsymbol{x})$ and $\sigma(x) = (1 + e^{-x})^{-1}$, representing the probability that the input sample is classified into the first class.
In multi-class LR classification (i.e., $c > 2$), a typical method is multinomial classification, which trains a linear regression model $\boldsymbol{\theta}^{(k)}$ for each class $k \in \{1, \cdots, c\}$ and applies a softmax function on the $c$ linear regression predictions $f(\boldsymbol{x}, \boldsymbol{\theta}^{(1)}), \cdots, f(\boldsymbol{x}, \boldsymbol{\theta}^{(c)})$. 
%
As a consequence, the output of the softmax function is the confidence score vector, i.e., the probabilities of the $c$ classes.

\vspace{1mm}\noindent
\textbf{Neural networks (NN) model prediction}.
The neural network model is a popular architecture in deep learning. In an NN model, $f$ is typically composed of an input layer, an output layer, and a sequence of hidden layers with non-linear transformations from the input to the output. 
The model parameters $\boldsymbol{\theta}$ denote the weights for each transformation \cite{SongRS17}.
In the prediction stage, the results of the output layer are the confidence scores.
%
%
%
%
%


\vspace{1mm}\noindent
\textbf{Tree-based model prediction.} 
The decision tree (DT) model is widely adopted in real-world applications due to their good interpretability.
The DT model consists of a number of internal nodes and leaf nodes.
Each internal node describes a branching threshold \textit{w.r.t.} a feature, while each leaf node represents a class label.
When predicting an input sample, a sequence of branching operations are executed based on comparisons of the sample's feature values and the thresholds, until a leaf node is reached. 
The prediction output is the class label on that leaf node. Note that the branching operations are deterministic in the DT model. Therefore, the confidence score for the predicted class is 1, while the confidence scores for other classes are 0.

We can use ensemble models to further improve predictive performance, e.g., random forest (RF). 
Essentially, the RF model is composed of a set of independent DT models.
In the prediction stage, each tree predicts a class label, and the final classification is calculated by majority voting from all trees.
As a result, the prediction output of the RF model also includes a vector of confidence scores, where each element $v_k$ of class $k \in \{1, \cdots, c\}$ is the fraction of trees that predict $k$.

\subsection{Vertical Federated Learning} \label{subsec:vertical-federated-learning}

Vertical FL enables an active party (e.g., the bank in Fig.~\ref{fig:vertical-fl-example}) to enrich his business data by incorporating more diverse features of users from one or several passive parties (e.g., the Fintech company in Fig~\ref{fig:vertical-fl-example}). 
From which the active party could improve his model accuracy to provide better customer service. Meanwhile, the passive parties could benefit from a pay-per-use model \cite{WuZPM16} for their contributions to the model training and model prediction stages. 
For example, the active party may pay the passive parties according to the importance of their provided features \cite{simcollaborative}. 

In the prediction stage of vertical FL, a model prediction is usually initiated by the active party who sends a prediction request to other parties, along with an input sample id (e.g., 100 or 102 in Fig.~\ref{fig:vertical-fl-example}). 
The passive parties then prepare the corresponding feature values of that sample, and all parties jointly run a protocol to compute the prediction output. 
The protocol can be secure enough to protect each party's sensitive data, such that only the prediction output is revealed to the active party for making further decisions, and no intermediate information during the computation is disclosed~\cite{WuCXCO20}.

\section{Problem Statement}\label{sec:problem-statement}
%

\subsection{System Model}\label{subsec:problem:system-model}
We consider a set of $m$ distributed parties (or data owners) $\{P_1, \cdots, P_m\}$ who train a vertical FL model by consolidating their respective datasets. 
After obtaining the trained vertical FL model parameters $\boldsymbol{\theta}$, the parties collaboratively make predictions on their joint prediction dataset $\mathcal{D}_{\text{pred}}=\{D_1, \cdots, D_m\}$.
Each row in the dataset corresponds to a sample, and each column corresponds to a feature. 
Let $n$ be the number of samples and $d_i$ be the number of features in ${D}_i$, where $i \in \{1, \cdots, m\}$. 
We denote $D_i = \{\boldsymbol{x}_{i}^{t}\}_{t=1}^{n}$ where $\boldsymbol{x}_{i}^{t}$ represents the $t$-th sample of $D_i$. 
%
In this paper, we consider the supervised classification task and denote the number of classes by $c$.
Table \ref{table:notations} summarizes the frequently used notations. 
\par

In vertical FL, the datasets $\{{D}_1, \cdots, D_m\}$ share the same sample ids but with disjoint subsets of features \cite{YangLCT19}. 
In particular, we assume that the parties have determined and aligned their common samples using private set intersection techniques \cite{Pinkas0SZ15, ChenLR17} without revealing any information about samples not in the intersection. 
%

\subsection{Threat Model}\label{subsec:problem:threat-model}

We consider the semi-honest model \cite{Mohassel17, WuWZLCL18, ChowdhuryW0MJ20} where every party follows the protocol exactly as specified, but may try to infer other parties' private information based on the messages received. 
Specifically, this paper focuses on the situation that the active party is the adversary. The active party may also collude with other passive parties to infer the private feature values of a set of target passive parties. 
The strongest notion is that $m-1$ parties collude (including the active party) to infer the feature values of the remaining passive party. 
Note that this is true when there are only two participating parties in the data collaboration.
%
In addition, we assume that the active party has no background information of the passive parties' data distribution. However, he may know the passive parties' feature names, types, and value ranges; this is reasonable because the active party often needs this information to justify the effectiveness of the trained model. 

\ifthenelse{\boolean{isrevision}}
{
}
{
We do not consider malicious behaviors in this paper. On the one hand, a malicious active party who deviates from the agreed protocol will obtain inaccurate model predictions, leading to a wrong decision. The risk might be too high compared to the gain of the inferred information.
On the other hand, a malicious passive party who provides useless or incorrect feature values may receive no rewards and lose trust in his collaborators. For example, the active party could apply some incentive methods to justify the contributions of those provided data~\cite{simcollaborative}. 
}

\begin{table}[t]
\centering
\caption{Summary of notations}
\begin{tabular}{  l  l }
\toprule
Notation & Description \\
\toprule
$m$ & the number of parties in vertical FL \\
$n$ & the number of samples in the prediction dataset \\
$d$ & the number of total features \\
$c$ & the number of classes in the classification \\
$\boldsymbol{x}$ & an input sample in the prediction stage \\
$\boldsymbol{v}$ & the prediction output, i.e., confidence scores of $\boldsymbol{x}$ \\
${P}_{\text{adv}}$ & the adversary (the active party and several passive parties) \\
${P}_{\text{target}}$ & the attack target (the remaining passive parties) \\
${d}_{\text{adv}}$ & the number of features hold by ${P}_{\text{adv}}$ \\
${d}_{\text{target}}$ & the number of features hold by ${P}_{\text{target}}$ \\
${\boldsymbol{x}}_{\text{adv}}$ & the feature values of $\boldsymbol{x}$ hold by ${P}_{\text{adv}}$ \\
${\boldsymbol{x}}_{\text{target}}$ & the feature values of $\boldsymbol{x}$ hold by ${P}_{\text{target}}$ \\
$\boldsymbol{\theta}$ & the parameters of the trained vertical FL model \\
$\boldsymbol{\theta}_G$ & the parameters of the generator model \\
\bottomrule
\end{tabular}
\label{table:notations}
\end{table}

\subsection{Feature Inference Attack on Model Predictions} \label{subsec:problem:vfl-prediction}

Given the vertical FL model parameters $\boldsymbol{\theta}$, the parties can jointly make predictions on new input samples, whose feature values are distributed among $m$ parties. 
As described in Section \ref{subsec:background:machine-learning}, the prediction output consists of a vector of confidence scores, i.e., $\boldsymbol{v} = (v_1, \cdots, v_c)$.
%
%
%
In this paper, we assume that $\boldsymbol{\theta}$ and $\boldsymbol{v}$ are revealed to the active party because this information is essential for the active party to make correct and interpretable decisions. 
\par

Let $\boldsymbol{x}$ be an input sample for prediction. 
Without loss of generality, the $m$ parties can be abstracted into two parties: the adversary $P_{\text{adv}}$, which is composed of the active party and a subset of colluding passive parties; and the attack target $P_{\text{target}}$, which is composed of the remaining passive parties. 
Let $\boldsymbol{x} = (\boldsymbol{x}_{\text{adv}}, \boldsymbol{x}_{\text{target}})$, such that $\boldsymbol{x}_{\text{adv}}$ and $\boldsymbol{x}_{\text{target}}$ denote the feature values held by $P_{\text{adv}}$ and $P_{\text{target}}$, respectively; and $d_{\text{adv}}$ and $d_{\text{target}}$ denote the number of features held by $P_{\text{adv}}$ and $P_{\text{target}}$ accordingly.

As a consequence, the setting for our feature inference attack is as follows. The adversary $P_{\text{adv}}$ is given the vertical FL model parameters $\boldsymbol{\theta}$, the prediction output $\boldsymbol{v}$, and the feature values $\boldsymbol{x}_{\text{adv}}$ that belongs to himself. $P_{\text{adv}}$'s goal is to infer the feature values of $P_{\text{target}}$, i.e., 
\begin{align}
    \label{eq:attack-problem}
    \hat{\boldsymbol{x}}_{\text{target}} = \mathcal{A}(\boldsymbol{x}_{\text{adv}}, \boldsymbol{v}, \boldsymbol{\theta})
\end{align}
where $\hat{\boldsymbol{x}}_{\text{target}}$ is the inferred feature values and $\mathcal{A}$ is an attack algorithm executed by $P_{\text{adv}}$. 
Specifically, we use two metrics to evaluate the attack performance. 
One is \textit{mean square error} (MSE), measuring the deviation of the inferred feature values from the ground-truth $\boldsymbol{x}_{\text{target}}$.
The other is \textit{correct branching rate} (CBR), which is defined for the attacks on tree-based models, measuring the fraction of inferred feature values that belong to the same branches as those computed by the ground-truth.
A detailed discussion will be presented in Section~\ref{sec:experiments}.
\section{Attack Based on Individual Prediction}
\label{sec:attacks-individual-prediction}

In this section, we consider the attack problem in Eqn~(\ref{eq:attack-problem}) based on individual model prediction, which means the adversary can initiate an attack once obtaining the prediction output of a sample. 
Specifically, we present an equality solving attack on the logistic regression (LR) model and a path restriction attack on the decision tree (DT) model in Section \ref{subsec:attacks-lr-nn:specfic} and \ref{subsec:attack-dt}, respectively.

\subsection{Equality Solving Attack}
\label{subsec:attacks-lr-nn:specfic}


As described in Section~\ref{subsec:background:machine-learning}, the LR model prediction of sample $\boldsymbol{x}$ is computed by a deterministic function $f(\boldsymbol{x}, \boldsymbol{\theta})$, where $\boldsymbol{\theta}$ is the model parameters known to the active party (i.e., the adversary). Therefore, given the prediction output $\boldsymbol{v}$, the adversary $P_{\text{adv}}$ can construct a set of equations, from which $P_{\text{adv}}$ could infer the feature values held by $P_{\text{target}}$. We discuss the binary LR classification and multi-class LR classification separately.

%

\vspace{1mm}\noindent 
\textbf{Binary LR prediction.} 
We denote the model parameters as $\boldsymbol{\theta} = (\boldsymbol{\theta}_{\text{adv}}, \boldsymbol{\theta}_{\text{target}})$, where $\boldsymbol{\theta}_{\text{adv}}$ and $\boldsymbol{\theta}_{\text{target}}$ are the weights corresponding to the features owned by $P_{\text{adv}}$ and  $P_{\text{target}}$, respectively.
Let $\boldsymbol{v}$ be the prediction output of a given sample $\boldsymbol{x} = (\boldsymbol{x}_{\text{adv}}, \boldsymbol{x}_{\text{target}})$.
%
Note that for binary LR classification, there is only one meaningful confidence score in $\boldsymbol{v}$, e.g., $v_1$ for the first class, and that for the second class can be derived by $1-v_1$.
Given $v_1$ and the adversary's own feature values $\boldsymbol{x}_{\text{adv}}$, it is straightforward for $P_{\text{adv}}$ to create an equation with $\boldsymbol{x}_{\text{target}}$ as the variables, i.e.,
\begin{align}\label{eq:single-lr-equation}
    \sigma(\boldsymbol{x}_{\text{adv}} \cdot \boldsymbol{\theta}_{\text{adv}} + \boldsymbol{x}_{\text{target}} \cdot \boldsymbol{\theta}_{\text{target}}) = v_1
\end{align}
where $\sigma(\cdot)$ is the sigmoid function. Obviously, if there is only one unknown feature, i.e., $d_{\text{target}} = 1$, then the equation has only one solution, which means the unknown feature value $\boldsymbol{x}_{\text{target}}$ can be inferred precisely.
\par

\vspace{1mm}\noindent
\textbf{Multi-class LR prediction.}
For multi-class LR classification, as mentioned in Section \ref{subsec:background:machine-learning}, there are $c$ linear regression models. Let $\boldsymbol{\theta} = (\boldsymbol{\theta}^{(1)}, \cdots, \boldsymbol{\theta}^{(c)})$ be the parameters for these $c$ models, respectively.
%
To initiate the feature inference attack, the adversary aims to construct the following linear equations for $k \in \{1,\cdots,c\}$.
\begin{align}\label{eq:lr-equation-system}
    \boldsymbol{x}_{\text{adv}} \cdot \boldsymbol{\theta}_{\text{adv}}^{(k)} & + \boldsymbol{x}_{\text{target}} \cdot \boldsymbol{\theta}_{\text{target}}^{(k)} = z_k 
\end{align}
where $z_k$ is the output of the $k$-th linear regression model.
However, $P_{\text{adv}}$ only knows the confidence score vector $\boldsymbol{v} = (v_1, \cdots, v_{c})$ such that 
\begin{align}
    \label{eq:softmax-function}
    v_k = \frac{\exp{(z_k)}}{\sum\nolimits_{k'}\exp{(z_{k'})}}.
\end{align}
Also, it is impossible for $P_{\text{adv}}$ to revert $z_1, \cdots, z_{c}$ from $v_1, \cdots, v_{c}$ because there are multiple solutions to Eqn~(\ref{eq:softmax-function}). 
Nevertheless, we observe that an important correlation exists between two confidence scores. Specifically, we can transform Eqn~(\ref{eq:softmax-function}) into:
\begin{align}
   \ln v_k & = z_k - \ln (\sum\nolimits_{k'} \exp(z_{k'})). \label{eq:softmax-conversion-2}
\end{align}
Note that the second term in the right hand of Eqn~(\ref{eq:softmax-conversion-2}) is the same for all $k \in \{1,\cdots,c\}$. As a result, $P_{\text{adv}}$ can subtract $\ln v_k$ by another confidence score, say $\ln v_{k'}$, obtaining:
\begin{align}
    \label{eq:subtraction}
    \ln v_k - \ln v_{k'} = z_k - z_{k'}.
\end{align}
This property allows $P_{\text{adv}}$ to obtain $c-1$ linear equations by subtracting two adjacent equations in Eqn~(\ref{eq:lr-equation-system}), i.e.,
\begin{align}
    \label{eq:converted-equations}
        \boldsymbol{x}_{\text{adv}} (\boldsymbol{\theta}_{\text{adv}}^{(k)} - \boldsymbol{\theta}_{\text{adv}}^{(k+1)}) & + \boldsymbol{x}_{\text{target}} (\boldsymbol{\theta}_{\text{target}}^{(k)} - \boldsymbol{\theta}_{\text{target}}^{(k+1)}) = a_k'
\end{align}
where $a_k' = \ln v_k - \ln v_{k+1}$ and $k \in \{1,\cdots, c-1\}$. 
Consequently, if the number of unknown features $d_{\text{target}} \leq c-1$, then there will be only one solution for $\boldsymbol{x}_{\text{target}}$, which can be inferred exactly. 

\vspace{1mm}\noindent
\textbf{Attack method.}
Based on the discussion for both binary LR and multi-class LR, we observe that when a specific condition is satisfied, i.e., $d_{\text{target}} \leq c-1$, $P_{\text{adv}}$ can infer $\boldsymbol{x}_{\text{target}}$ precisely. 
Due to that Eqn~(\ref{eq:single-lr-equation}) and Eqn~(\ref{eq:converted-equations}) are linear equations, we can rewrite them into matrix form, i.e., $\boldsymbol{\Theta}_{\text{target}} \boldsymbol{x}_{\text{target}} = \boldsymbol{a}$, where the dimension of $\boldsymbol{\Theta}_{\text{target}}$ is $(c-1) \times d_{\text{target}}$, and $\boldsymbol{x}_{\text{target}}$ and $\boldsymbol{a}$ are column vectors with size $d_{\text{target}}$ and $c-1$, respectively.
In particular, for binary LR, $\boldsymbol{\Theta}_{\text{target}}$ equals to $\boldsymbol{\theta}_{\text{target}}$ and $\boldsymbol{a}$ is $\sigma^{-1}(v_1) - \boldsymbol{x}_{\text{adv}} \cdot \boldsymbol{\theta}_{\text{adv}}$. While for multi-class LR, $\boldsymbol{\Theta}_{\text{target}}$ is composed of $c-1$ vectors $\boldsymbol{\theta}_{\text{target}}^{(k)} - \boldsymbol{\theta}_{\text{target}}^{(k+1)}$ for $k \in \{1, \cdots, c-1\}$ and $\boldsymbol{a}$ is derived from Eqn~(\ref{eq:converted-equations}) by substituting the known values. 
Given this matrix representation, the adversary can solve the target feature values easily by $\hat{\boldsymbol{x}}_{\text{target}} = {\boldsymbol{\Theta}}_{\text{target}}^{+} \boldsymbol{a}$, where ${\boldsymbol{\Theta}}_{\text{target}}^{+}$ is the pseudo-inverse matrix \cite{pseudoinverse} of $\boldsymbol{\Theta}_{\text{target}}$. 
%

When $d_{\text{target}} \geq c$, although there are infinite solutions to Eqn~(\ref{eq:converted-equations}), the solution $\hat{\boldsymbol{x}}_{\text{target}}$ constructed by ${\boldsymbol{\Theta}}_{\text{target}}^{+} \boldsymbol{a}$, which minimizes $||\boldsymbol{\Theta}_{\text{target}} \hat{\boldsymbol{x}}_{\text{target}} - \boldsymbol{a}||_2$ and satisfies $||\hat{\boldsymbol{x}}_{\text{target}}||_2\leq ||\boldsymbol{x}_{\text{target}}||_2$ for all solutions (we refer the interested readers to~\cite{pseudoinverse} for more details), can still give a good estimation of $\boldsymbol{x}_{\text{target}}$.
We will experimentally demonstrate this in Section~\ref{sec:experiments}.

\newtheorem{example}{Example}
\begin{example}\label{example:esa}
Suppose the Bank in Fig.~\ref{fig:vertical-fl-example} tries to infer a user's unknown feature values that belong to the FinTech company. To illustrate the equality solving attack, we assume that there are 3 classes and the trained model parameters are $\boldsymbol{\Theta} = \begin{bmatrix}
\small
\boldsymbol{\theta}^{(1)}\\
\boldsymbol{\theta}^{(2)}\\
\boldsymbol{\theta}^{(3)}
\end{bmatrix} = 
\begin{bmatrix}
\small
0.08 & 0.0002 & 0.0005 & 0.09\\
0.06 & 0.0005 & 0.0002 & 0.08\\
0.01 & 0.0001 & 0.0004 & 0.05
\end{bmatrix}$. Suppose there is an input sample $\boldsymbol{x} = (25, 2\text{K}, 8\text{K}, 3)$, and the predicted confidence score vector is $\boldsymbol{v} = (0.867, 0.084, 0.049)$. $P_{\text{adv}}$ has the former two feature values `age=25' and `income=2K'. Then, from Eqn~(\ref{eq:subtraction})-(\ref{eq:converted-equations}), he can compute $\boldsymbol{\theta}_{\text{adv}}^{(1)} - \boldsymbol{\theta}_{\text{adv}}^{(2)}= (0.02, -0.0003)$, $\boldsymbol{\theta}_{\text{adv}}^{(2)} - \boldsymbol{\theta}_{\text{adv}}^{(3)}= (0.05, 0.0004)$, $\boldsymbol{\theta}_{\text{target}}^{(1)} - \boldsymbol{\theta}_{\text{target}}^{(2)}= (0.0003, 0.01)$, $\boldsymbol{\theta}_{\text{target}}^{(2)} - \boldsymbol{\theta}_{\text{target}}^{(3)}= (-0.0002, 0.03)$, $a_1' = 2.334$, and $a_2' = 0.539$. Consequently, $P_{\text{adv}}$ can construct two equations and estimate $\hat{\boldsymbol{x}}_{\text{target}} = (8011.8, 3.046)$, where the loss is from the precision truncation during the computations.
\end{example}

\subsection{Path Restriction Attack}
\label{subsec:attack-dt}
For the DT model, the prediction output includes only the predicted class, with a confidence score of 1.
We propose a specific attack called \textit{path restriction attack}, by which the adversary can restrict the possible prediction paths in the tree model based on the predicted class and his own feature values. 

\begin{figure}[t]
\centering
\includegraphics[width=\columnwidth]{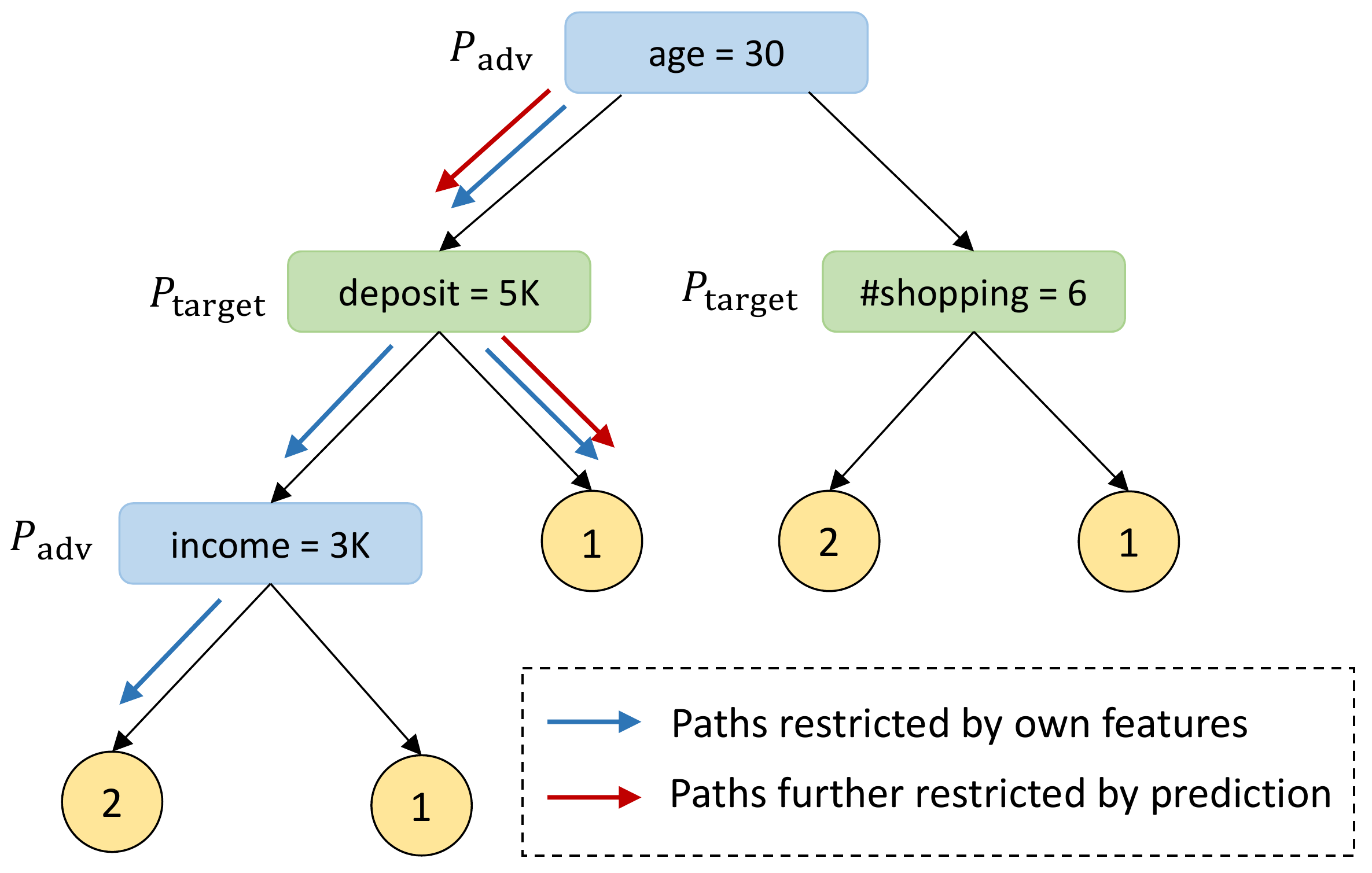}
\caption{Example of path restriction attack. Assume that $P_{\text{adv}}$'s feature values of an input sample is: `age=25' and `income=2K', while the predicted class is 1.}
\label{fig:path-restriction}
\end{figure}

\begin{example}\label{example:pra}
Fig.~\ref{fig:path-restriction} illustrates the basic idea of this attack given Fig.~\ref{fig:vertical-fl-example}. The adversary $P_{\text{adv}}$ has partial feature values: `age=25' and `income=2K'. Then $P_{\text{adv}}$ can restrict the possible prediction paths from 5 (the total number of prediction paths) to 2 (with blue arrows).
Besides, if assuming that the predicted class of this sample is 1, $P_{\text{adv}}$ can identify the real prediction path (with red arrows) and correctly infer that $P_{\text{target}}$'s deposit feature value of this sample is larger than 5K. 
\end{example}

Generally, let $n_p$ be the total number of prediction paths, the adversary can restrict the candidate prediction paths to $n_r$ paths by comparing his own feature values with the branching thresholds in the tree model and checking if the leaf label of a path matches the ground-truth predicted class. Consequently, for the inference, the adversary uniformly and randomly selects a path from the $n_r$ paths and checks the unknown feature values belong to which branch compared to the branching thresholds.

\begin{algorithm}[t]
\DontPrintSemicolon
\small
\KwIn{${\boldsymbol{x}_{\text{adv}}}$: adversary's feature values,
${\boldsymbol{\theta}}$: model parameters, 
$k$: predicted class
}
\KwOut{
$\mathcal{P}$: possible prediction paths
}
{
    $\boldsymbol{\beta} = \boldsymbol{0}$ $\slash\slash$ initialize indicator vector  \\
    $Q \leftarrow $ root node (index 0) \\
    $\beta_0 = 1$ \\
    \While{$Q$ is not empty} {
        node $i$ $\leftarrow$ $Q.\textbf{pop}()$ \\
        \If{node $i$ belongs to the adversary} {
            \If{adversary's feature value $\leq$ threshold value} {
                $\beta_{2i+1} = \beta_i$, $\beta_{2i+2} = 0$
            }
            \Else{
                $\beta_{2i+1} = 0$, $\beta_{2i+2} = \beta_i$
            }
        }
        \Else{
            $\beta_{2i+1} = \beta_{2i+2} = \beta_{i}$
        }
        \If{node $i$ is not leaf node}{
            $Q.\textbf{push}(2i+1)$, $Q.\textbf{push}(2i+2)$ \\
        }
    }
    $\boldsymbol{\alpha} \leftarrow$ compute the indicator vector, such that the elements of leaf nodes with label $k$ are set to 1 and otherwise 0 \\ 
    $\boldsymbol{\beta} \leftarrow$ element-wise multiplication of $\boldsymbol{\alpha}$ and $\boldsymbol{\beta}$ \\
    $\mathcal{P} \leftarrow$ find the paths that $\beta' = 1$ ($\forall \beta' \in \boldsymbol{\beta}$) \\
    return $\mathcal{P}$
}
\caption{Path restriction attack}\label{alg:path-restriction}
\end{algorithm}


Algorithm \ref{alg:path-restriction} summarizes this attack method. 
Let $n_{f}$ be the number of tree nodes in the full binary tree. 
The adversary $P_{\text{adv}}$ first initializes an indicator vector $\boldsymbol{\beta}$ of size $n_f$ with 0 (line 1).
This vector is used for recording the tree nodes that may be evaluated in the perspective of $P_{\text{adv}}$.
Then $P_{\text{adv}}$ initializes a queue $Q$ with the root node (whose index is 0) and sets $\beta_0 = 1$ because the root node must be evaluated for any prediction (lines 2-3).
Next, $P_{\text{adv}}$ iteratively pops a node from $Q$ and checks the following conditions until $Q$ is empty (lines 4-14). 
If the feature on the current node belongs to $P_{\text{adv}}$, $P_{\text{adv}}$ updates the corresponding indicators of its child nodes in $\boldsymbol{\beta}$ according to the comparison result between his feature value and the branching threshold on that node (lines 6-10).
Otherwise, the indicators of the child nodes are the same as the indicator of the current node (line 12).
For example, in Fig.~\ref{fig:path-restriction}, given the branching condition `age=30' on the root node ($i = 0$) and the feature value `age=25' of the input sample, $P_{\text{adv}}$ can update $\beta_{2i+1} = \beta_i = 1$ and $\beta_{2i+2} = 0$ as the prediction goes into the left branch. 
After that, if the current node is not a leaf node, $P_{\text{adv}}$ pushes its child nodes into $Q$ for iteratively updating. 
As a result, $\boldsymbol{\beta}$ can be computed. For example, $\boldsymbol{\beta} = (1,1,0,1,1,0,0,1,0,0,0,0,0,0,0)$ given Fig.~\ref{fig:path-restriction}.
In the following, $P_{\text{adv}}$ computes another indicator vector $\boldsymbol{\alpha}$ with size $n_f$, such that the elements of leaf nodes with label $k$ are set to 1 and the others are set to 0 (line 15). 
For example, $\boldsymbol{\alpha} = (0,0,0,0,1,0,1,0,1,0,0,0,0,0,0)$ in Fig.~\ref{fig:path-restriction}.
Finally, $P_{\text{adv}}$ updates $\boldsymbol{\beta}$ by element-wise multiplication using $\boldsymbol{\alpha}$, and the restricted prediction paths are those elements with 1 in $\boldsymbol{\beta}$ (lines 16-17).
In Fig.~\ref{fig:path-restriction}, there is only one element is 1 (i.e., node index 4) in the updated $\boldsymbol{\beta}$, and the corresponding prediction path can be identified (i.e., with red arrows).
%
The complexity of Algorithm~\ref{alg:path-restriction} is $O(n_f)$ as the adversary needs to traverse the tree model to compute the indicator vectors.


\section{Attack Based on Multiple Predictions}\label{sec:attack-multiple-predictions}

So far, we identify two specific attacks based on individual model prediction. However, these attacks could hardly be applied to complex models, such as neural networks (NN) and random forest (RF).
%
For the NN model, there are a number of hidden layers for non-linear transformations, making it difficult to solve the equations by the equality solving attack. Even though the adversary could utilize the maximum a posteriori (MAP) method (e.g., \cite{fredrikson2015model}) to find a plausible solution, the attack accuracy would be undesirable because the solution space to the unknown features in the NN model is huge and irregular.
For the RF model, given the prediction output (i.e., the confidence scores), the number of candidate tree combinations might be too large. For example, if there are 100 trees in the RF model and a prediction output with two classes is $\boldsymbol{v} = (0.4,0.6)$, then the adversary needs to consider $C_{100}^{40}$ combinations of the trees, which is computationally expensive.



To address the above limitations, we design a general feature inference attack, called \textit{generative regression network} (GRN), based on multiple model predictions. 
The rationale that the adversary can rely on multiple predictions is that the active party can easily collect this information by observing model predictions of new samples in the long term, for example, in a week or a month, as long as the vertical FL model is unchanged.
These accumulated data could be used for initiating a more sophisticated inference attack. 

\begin{figure*}[t]
\centering
\includegraphics[width=0.75\textwidth]{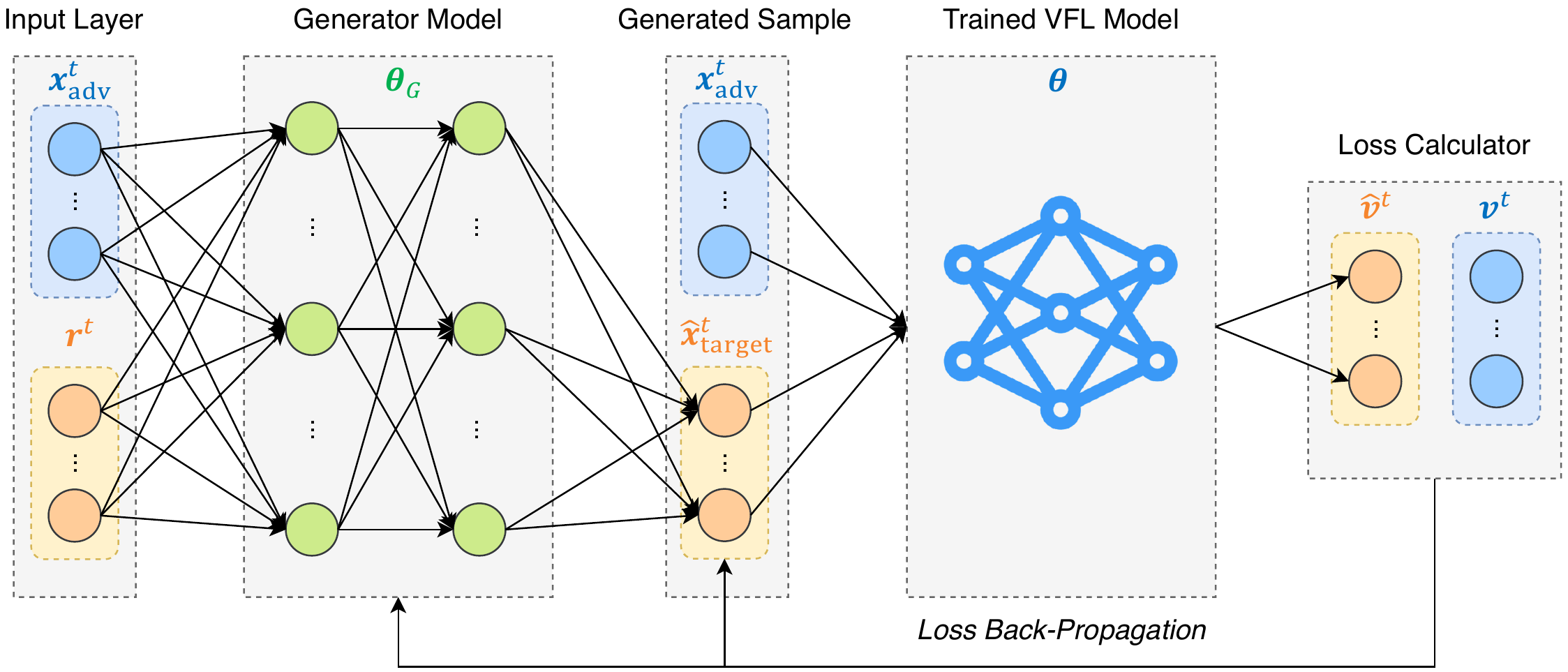}
\caption{Illustration of generative regression network trained by the adversary
}
\vspace{-0.5cm}
\label{fig:generator-attack-illustration}
\end{figure*}

Fig.~\ref{fig:generator-attack-illustration} gives an overview of this attack. The basic idea is to figure out the overall correlations between the adversary's own features and the attack target's unknown features. 
Upon this, the problem of inferring the unknown feature values is equivalent to the problem of generating new values $\hat{\boldsymbol{x}}_{\text{target}}$ to match the decisions of vertical FL model, where $\hat{\boldsymbol{x}}_{\text{target}}$ follows a probability distribution determined by the adversary's known feature values and the feature correlations.
To learn such a probability distribution, we build a generator model (in green color in Fig.~\ref{fig:generator-attack-illustration}) which takes the adversary's known feature values and a set of random variables as inputs (as depicted in the input layer in Fig.~\ref{fig:generator-attack-illustration}) and produces an estimation of the unknown feature values. The estimated values and the known feature values are combined into a \textit{generated sample}. 
Consequently, the generator model is trained by minimizing the loss between the predictions of the generated samples and the ground-truth predictions.
%


\begin{example}\label{example:grna}
Suppose that the Bank in Fig.~\ref{fig:vertical-fl-example} knows feature values $\boldsymbol{x}^t_{\text{adv}}=(\text{age }25, \text{income }2\text{K})$, and the prediction output $v^t=0.7$ if there are two classes. 
To infer the other two feature values, the Bank first feeds $\boldsymbol{x}^t_{\text{adv}}$ together with some random values $(25, 2\text{K}, r^t_1, r^t_2)$ into the generator to obtain a preliminary result, e.g. $\hat{\boldsymbol{x}}^t_{\text{target}}=(2\text{K}, 2)$;
then, $\boldsymbol{x}^t_{\text{adv}}$ and  $\hat{\boldsymbol{x}}^t_{\text{target}}$ are concatenated and input into the federated model to get a simulated output $\hat{v}^t=0.5$; 
the loss function $\ell(\hat{v}^t, v^t)$ is then calculated and back-propagated to the generator. 
With multiple predictions, the generator can learn the data distribution of the target features. 
\end{example}

Note that the random vector $\boldsymbol{r}^t$ is an indispensable component of GRN. First, it acts as a regularizer, encouraging the generator to capture more randomness arising from $\boldsymbol{x}^t_{\text{adv}}$. Second, in different epochs, the $\boldsymbol{x}^t_{\text{adv}}$ of a specific sample can be concatenated with different $\boldsymbol{r}^t$ into different generator inputs, which derive various gradient directions during the back-propagation and thus lead to a better estimation of $\boldsymbol{x}^t_{\text{target}}$ with high probabilities. 
In Section~\ref{subsec:experiments:multiple}, we show that a generator with random inputs can reduce the reconstruction error by $20\%$ compared to those without random inputs.

Unlike existing feature inference attacks, our attack method considers the most stringent case: it relies on no intermediate information disclosed during the computation (required in~\cite{MelisSCS19, ZhuLH19, hitaj2017deep, wang2019beyond}) and no background information of the attack target's data distribution (such as statistics or marginal feature distribution required in~\cite{fredrikson2014privacy, fredrikson2015model, yeom2018privacy}). 
In particular, the adversary only needs a set of model predictions and the vertical FL model for initiating the attack. 
In this section, we first present this attack on the NN model in Section~\ref{subsec:attacks-grn} and then describe how to extend it to the RF model in Section~\ref{subsec:attack-rf}.

\subsection{Generative Regression Network}
\label{subsec:attacks-grn}



We assume that the adversary $P_{\text{adv}}$ has collected prediction outputs of $n$ samples in $\mathcal{D}_{\text{pred}}$, where $\mathcal{D}_{\text{pred}} = (D_{\text{adv}}, D_{\text{target}})$ such that the $t$-th ($t \in \{1, \cdots, n\}$) sample is represented by $\boldsymbol{x}^{t} = (\boldsymbol{x}_{\text{adv}}^{t}, \boldsymbol{x}_{\text{target}}^{t})$. 
Let $V = (\boldsymbol{v}^1, \cdots, \boldsymbol{v}^{n})$ be the corresponding $n$ prediction outputs. 
The adversary's objective is to train a generator model, say $\boldsymbol{\theta}_G$, such that given the known feature values of the $t$-th sample $\boldsymbol{x}_{\text{adv}}^t$ and a random value vector $\boldsymbol{r}^t$ with size $d_{\text{target}}$, the generator outputs the corresponding estimation $\hat{\boldsymbol{x}}_{\text{target}}^t$ of the target party's $\boldsymbol{x}_{\text{target}}^t$.
To train the model, $P_{\text{adv}}$ can apply the mini-batch stochastic gradient descent method, where the training dataset is composed of $D_{\text{adv}}$ and $V$. 
The objective function is as follows:
\begin{align}
    \label{eq:generator-obj-func}
    \min_{\boldsymbol{\theta}_G} \frac{1}{n} \sum_{t=1}^{n} \ell (f(\boldsymbol{x}_{\text{adv}}^t, f_G(\boldsymbol{x}_{\text{adv}}^t, \boldsymbol{r}^t; \boldsymbol{\theta}_G); \boldsymbol{\theta}), \boldsymbol{v}^t) + \Omega(f_G)
\end{align}
where $\boldsymbol{\theta}$ and $\boldsymbol{\theta}_G$ are the parameters of the vertical FL model and the generator model, respectively. 
Moreover, $f_G$ denotes $\hat{\boldsymbol{x}}_{\text{target}}^t$, i.e., the output of the generator, and $f$ denotes the output of the vertical FL model given the generated sample (concatenated by $\boldsymbol{x}_{\text{adv}}^t$ and $\hat{\boldsymbol{x}}_{\text{target}}^t$). 
Besides, $\Omega(\cdot)$ is a regularization term of the generated unknown feature values $\{\hat{\boldsymbol{x}}_{\text{target}}^t\}_{t=1}^n$. For example, we penalize the generator model when the variance of $\{\hat{\boldsymbol{x}}_{\text{target}}^t\}_{t=1}^n$ is too large, preventing from generating meaningless samples. Nevertheless, the variance is computed based on the generated values, thus no prior information is needed by the adversary.

Algorithm~\ref{alg:generator-attack} presents the training method for GRN.
Specifically, in each iteration of each epoch, $P_{\text{adv}}$ first selects a batch of samples $B = (\boldsymbol{x}_{\text{adv}}^1, \cdots, \boldsymbol{x}_{\text{adv}}^{|B|})$ from $D_{\text{adv}}$ and initializes a set of $|B|$ random vectors $R = (\boldsymbol{r}^1, \cdots, \boldsymbol{r}^{|B|})$. The size of each random vector is the number of features held by the attack target $P_{\text{target}}$.
Then for the $t$-th sample in $B$, $P_{\text{adv}}$ feeds $\boldsymbol{x}_{\text{adv}}^t$ along with the corresponding $\boldsymbol{r}^t$ into the generator, obtaining $\hat{\boldsymbol{x}}_{\text{target}}^t$.
Next, $P_{\text{adv}}$ concats $\boldsymbol{x}_{\text{adv}}^t$ with $\hat{\boldsymbol{x}}_{\text{target}}^t$ to obtain a complete generated sample and makes a prediction using the vertical FL model, resulting in a prediction output $\hat{\boldsymbol{v}}^t$. 
As a result, the loss of this generated sample can be calculated by a loss function $\ell(\cdot, \cdot)$, e.g., MSE, with the ground-truth prediction output $\boldsymbol{v}^t$.
After obtaining the losses of all samples in $B$, the adversary back-propagates the aggregated loss to update the parameters of the generator model $\boldsymbol{\theta}_G$. 
Finally, the adversary can obtain the trained generator model after all the training epochs.


After obtaining $\boldsymbol{\theta}_G$, the adversary can use it to infer the unknown feature values $D_{\text{target}}$ in $\mathcal{D}_{\text{pred}}$. Specifically, for any sample $\boldsymbol{x}$ in $\mathcal{D}_{\text{pred}}$, the adversary generates a random vector $\boldsymbol{r}$ and directly computes $f_G(\boldsymbol{x}_{\text{adv}} \cup \boldsymbol{r}, \boldsymbol{\theta}_G)$ as the inferred feature values.
Notice that the samples to be attacked are exactly the samples for training the generator model.
This property ensures that the adversary can accumulate as many model predictions as possible to train the generator and improve the attack performance. We will experimentally show the effect of the number of model predictions in Section~\ref{sec:experiments}.



\begin{figure}[b]
\centering
\includegraphics[width=.7\columnwidth]{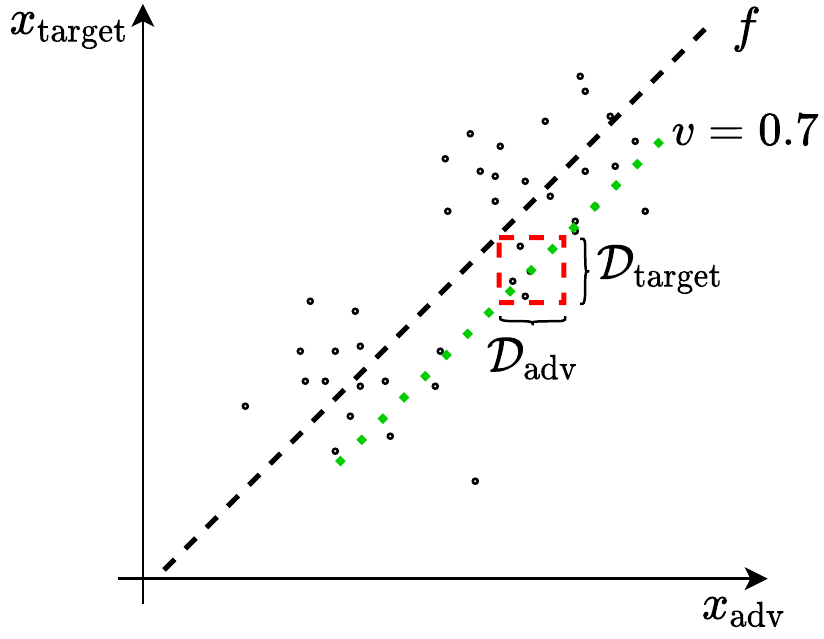}
\caption{An intuitive explanation for Generative Regression Network attack}
\label{fig:grn-explain}
\end{figure}

Fig.~\ref{fig:grn-explain} gives an intuitive explanation for the GRN attack. Suppose two parties collaboratively train a linear model $v=f(\theta_{\text{adv}} x_{\text{adv}}+ \theta_{\text{target}} x_{\text{target}})$. The black dashed line represents the decision boundary of $f$, and the green dashed line represents the domain of $\boldsymbol{x}=(x_{\text{adv}}, x_{\text{target}}) $ with confidence $v=0.7$. In the prediction 
stage, the adversary wishes to reconstruct the values of $x_{\text{target}}$ under $v=0.7$ based on the features $x_{\text{adv}}$ owned by himself and the model $f$. With $f$ and $v$, the possible domain of $x_{\text{target}}$ is reduced to the green dashed line. Then, according to the distribution of $x_{\text{adv}}$, GRN can infer the corresponding distribution of $x_{\text{target}}$ with high accuracy.

The GRN attack method is general because it takes the trained vertical FL model as a black-box, as long as that model's objective function is differentiable, such that the prediction loss could be back-propagated to the generator model. Thus, this attack works for both LR and NN models.

\subsection{Adopt GRN Attack on the Random Forest Model}
\label{subsec:attack-rf}

Let $W$ be the number of trees in the RF model. When predicting a sample, each tree produces a predicted class. The prediction output $\boldsymbol{v} = (v_1, \cdots, v_c)$ represents the fraction of trees that predict a label for each class. 
As discussed at the beginning of this section, the path restriction attack (see Section~\ref{subsec:attack-dt}) does not apply to the RF model, especially when $W$ is large.
Therefore, we desire to apply the GRN attack method on the RF model. However, since the objective function of the RF model is not differentiable, it is impossible to back-propagate the prediction loss through the RF model to the generator. 

To address this issue, we add an additional step in the attack method. After obtaining the vertical FL model (i.e., RF), the adversary can train another differentiable model (e.g., NN) to approximate the RF model~\cite{biau2019neural}. Specifically, the adversary first generates a number of dummy samples, say $D_{\text{dummy}}$ from the whole data space, then predicts each dummy sample by the RF model. Let $V_{\text{dummy}}$ be the prediction outputs. After that, the adversary could train an NN model $\boldsymbol{\theta}_{A}$ based on $(D_{\text{dummy}}, V_{\text{dummy}})$. 
Essentially, the NN model $\boldsymbol{\theta}_{A}$ is used to simulate the behavior of the RF model. 
As a consequence, the adversary can replace the RF model $\boldsymbol{\theta}$ with the new NN model $\boldsymbol{\theta}_{A}$ in Algorithm~\ref{alg:generator-attack} to train the generator model and infer the unknown feature values.

\begin{algorithm}[t]
\DontPrintSemicolon
\small
\KwIn{${\{\boldsymbol{x}_{\text{adv}}^t\}_{t=1}^{n}}$: adversary's feature values,
${\boldsymbol{\theta}}$: parameters of the vertical FL model, 
${\{\boldsymbol{v}^t\}}_{t=1}^n$: confidence scores, $\alpha$: learning rate
}
\KwOut{
$\boldsymbol{\theta}_G^*$: parameters of the generator model
}
{
    $\boldsymbol{\theta}_G \leftarrow \mathcal{N}(0,1)$ $\slash\slash$initialize generator model parameters \\
    \For{each epoch} {
        \For{each batch}{
            $loss = 0$ \\
            $B \leftarrow$ randomly select a batch of samples \\
            ${R} \leftarrow \mathcal{N}(0,1)$ $\slash\slash$initialize batch random values \\
            \For{$t \in \{1, \cdots, |B|\}$}{      $\hat{\boldsymbol{x}}_{\text{target}}^t \leftarrow f_G(\boldsymbol{x}_{\text{adv}}^t \cup \boldsymbol{r}^t; \boldsymbol{\theta}_G)$ \\ 
            $\hat{\boldsymbol{v}}^t \leftarrow f(\boldsymbol{x}_{\text{adv}}^t \cup \hat{\boldsymbol{x}}_{\text{target}}^t; \boldsymbol{\theta})$ \\
            $loss$ += $\ell(\hat{\boldsymbol{v}}^t, {\boldsymbol{v}}^t)$ \\
            }
            $\boldsymbol{\theta}_G \leftarrow \boldsymbol{\theta}_G - \alpha \cdot \bigtriangledown_{\boldsymbol{\theta}_G} loss$ $\slash\slash$update parameters
        }
    }
    return $\boldsymbol{\theta}_G$
}
\caption{Generative regression network training}\label{alg:generator-attack}
\end{algorithm}
\section{Experimental Evaluation}\label{sec:experiments}

This section shows the experimental evaluation of the proposed attacks. Section~\ref{subsec:experiments:setup} presents the experimental setup. The evaluations of the attacks based on individual model prediction and multiple model predictions are described in
Section~\ref{subsec:experiments:individual} and~\ref{subsec:experiments:multiple}, respectively.

\begin{figure*}[t]
\begin{subfigure}[b]{.49\columnwidth}
  \centering
  \includegraphics[width=0.95\columnwidth]{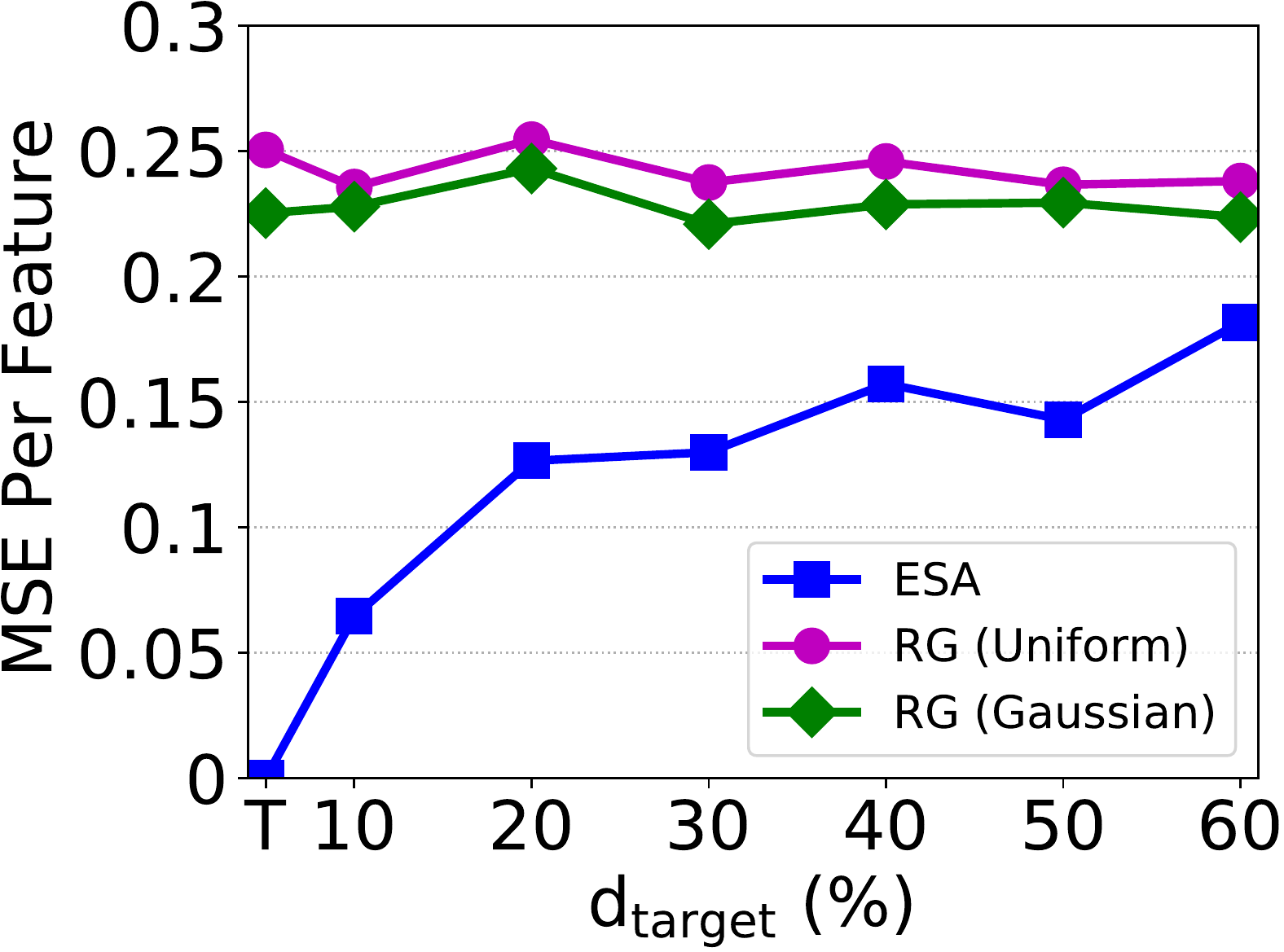}
  \caption{Bank marketing}
  \label{subfig:esa-bank}
\end{subfigure}
~
\begin{subfigure}[b]{.49\columnwidth}
  \centering
  \includegraphics[width=0.95\columnwidth]{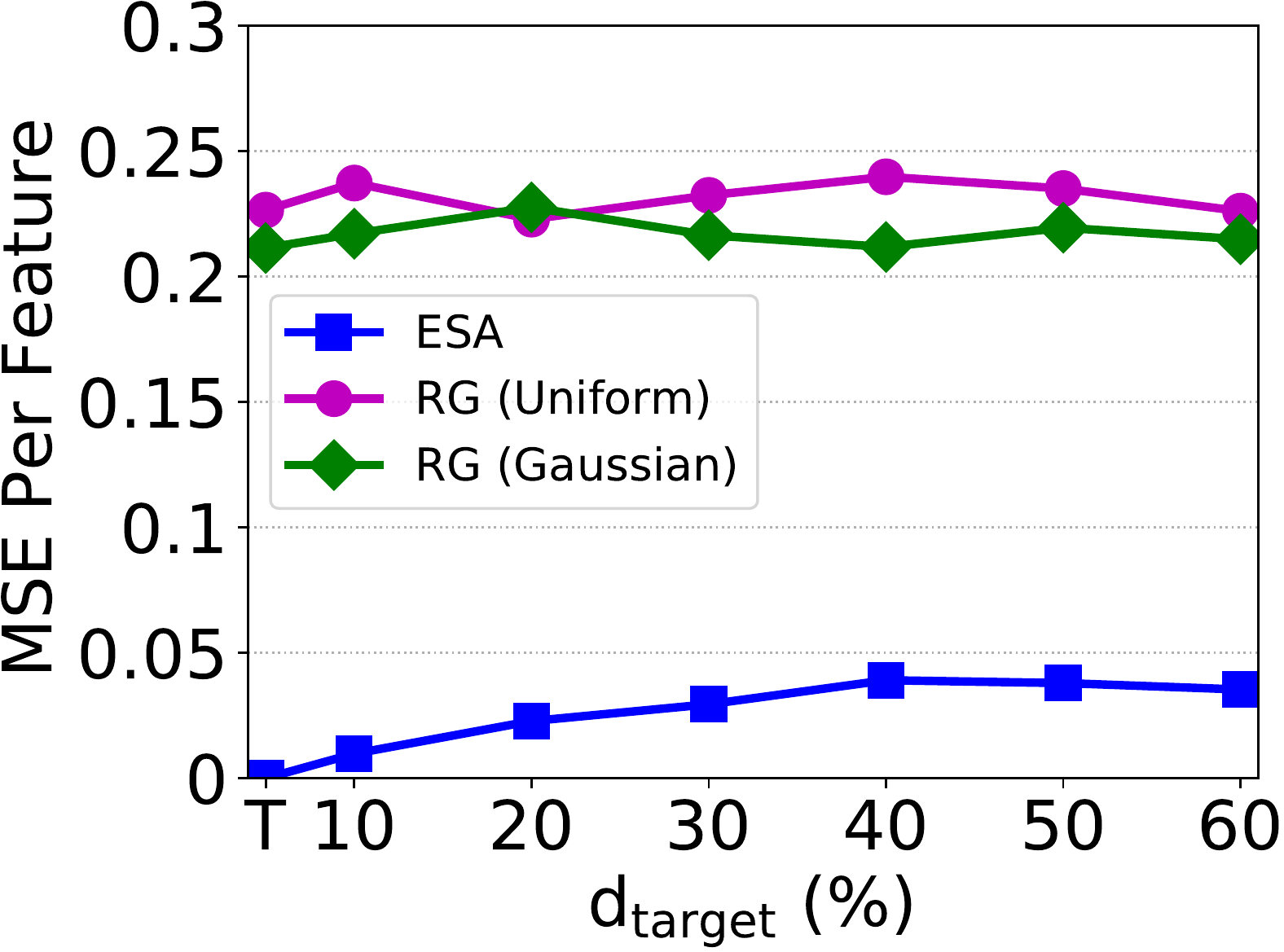}
  \caption{Credit card}
  \label{subfig:esa-credit}
\end{subfigure}
~
\begin{subfigure}[b]{.49\columnwidth}
  \centering
  \includegraphics[width=0.95\columnwidth]{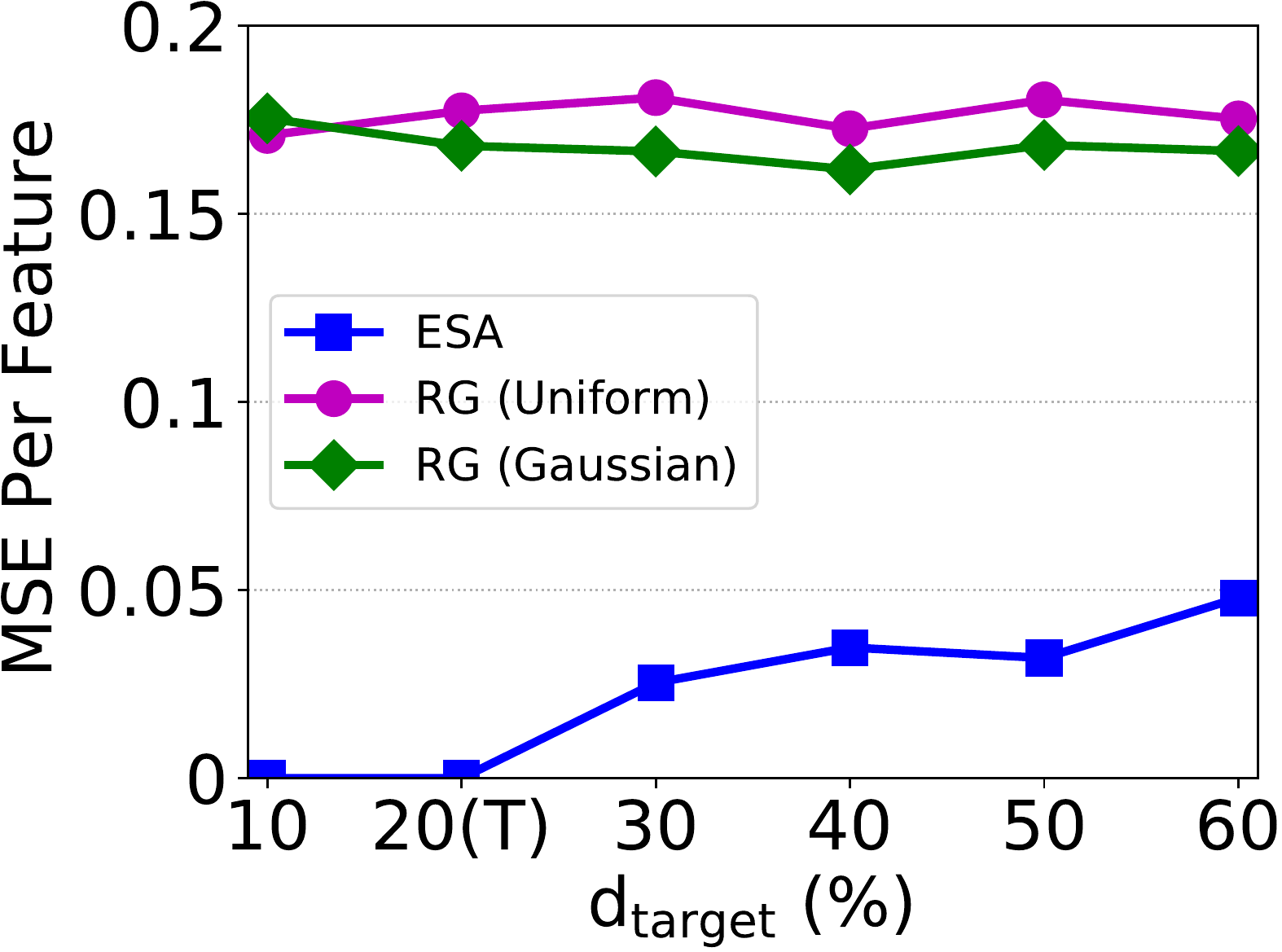}
  \caption{Drive diagnosis}
  \label{subfig:esa-drive}
\end{subfigure}
~
\begin{subfigure}[b]{.49\columnwidth}
  \centering
  \includegraphics[width=0.95\columnwidth]{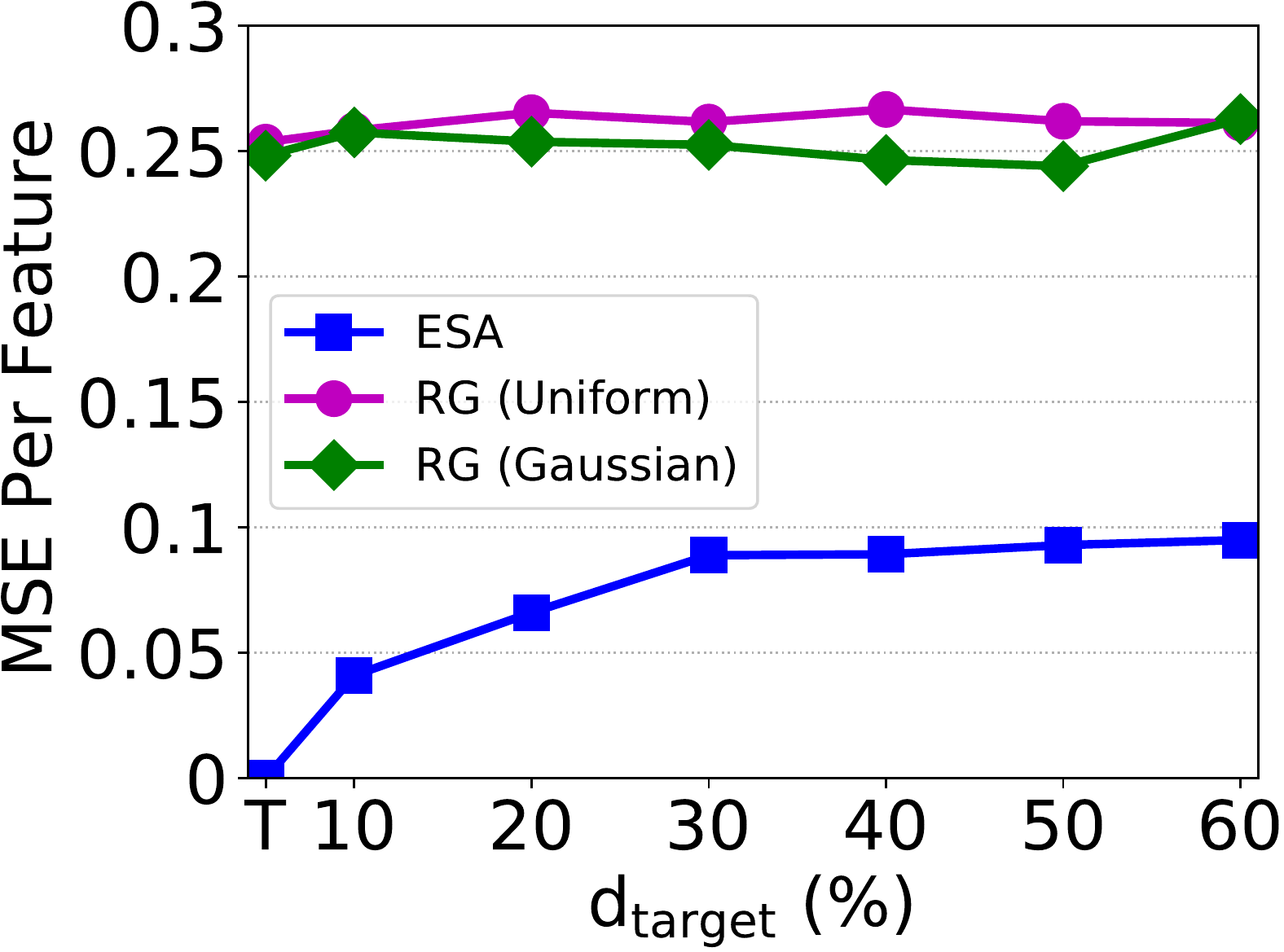}
  \caption{News popularity}
  \label{subfig:esa-news}
\end{subfigure}

\caption{Evaluation of equality solving attack \textit{w.r.t.} MSE per feature}
\vspace{-0.3cm}
\label{fig:esa-performance}
\end{figure*}

\begin{figure*}[t]
\begin{subfigure}[b]{.49\columnwidth}
  \centering
  \includegraphics[width=0.95\columnwidth]{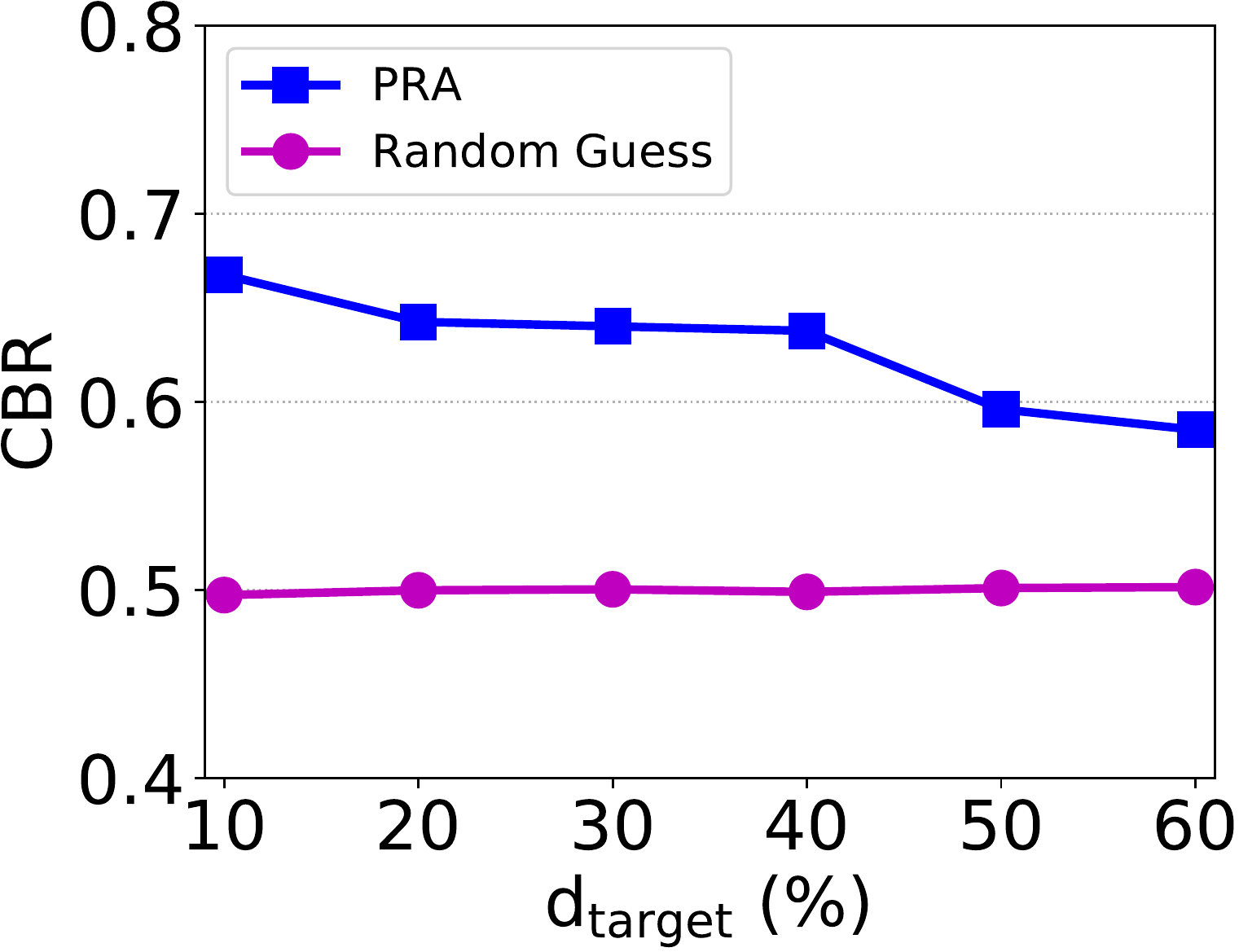}
  \caption{Bank marketing}
  \label{subfig:pra-bank}
\end{subfigure}
~
\begin{subfigure}[b]{.49\columnwidth}
  \centering
  \includegraphics[width=0.95\columnwidth]{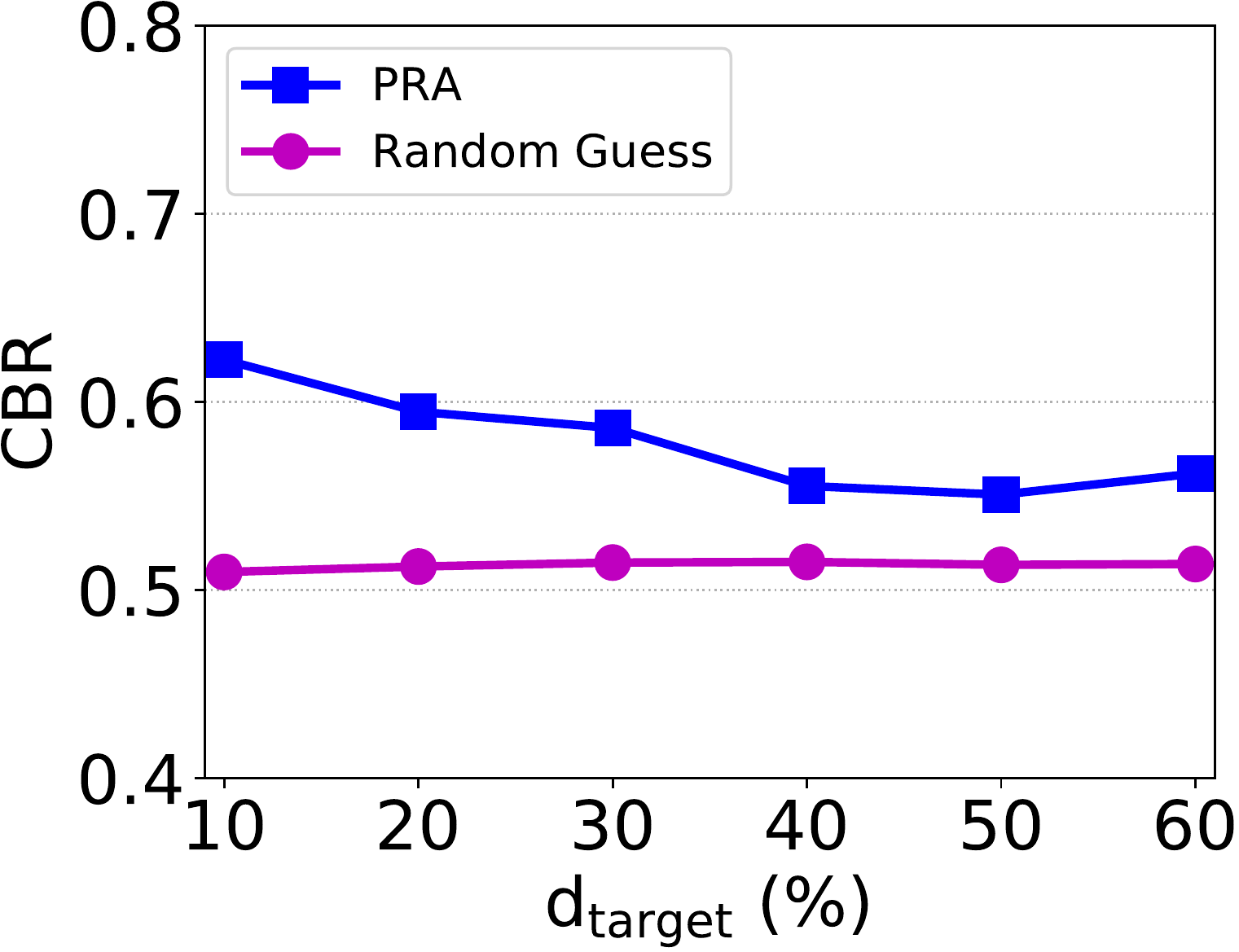}
  \caption{Credit card}
  \label{subfig:pra-credit}
\end{subfigure}
~
\begin{subfigure}[b]{.49\columnwidth}
  \centering
  \includegraphics[width=0.95\columnwidth]{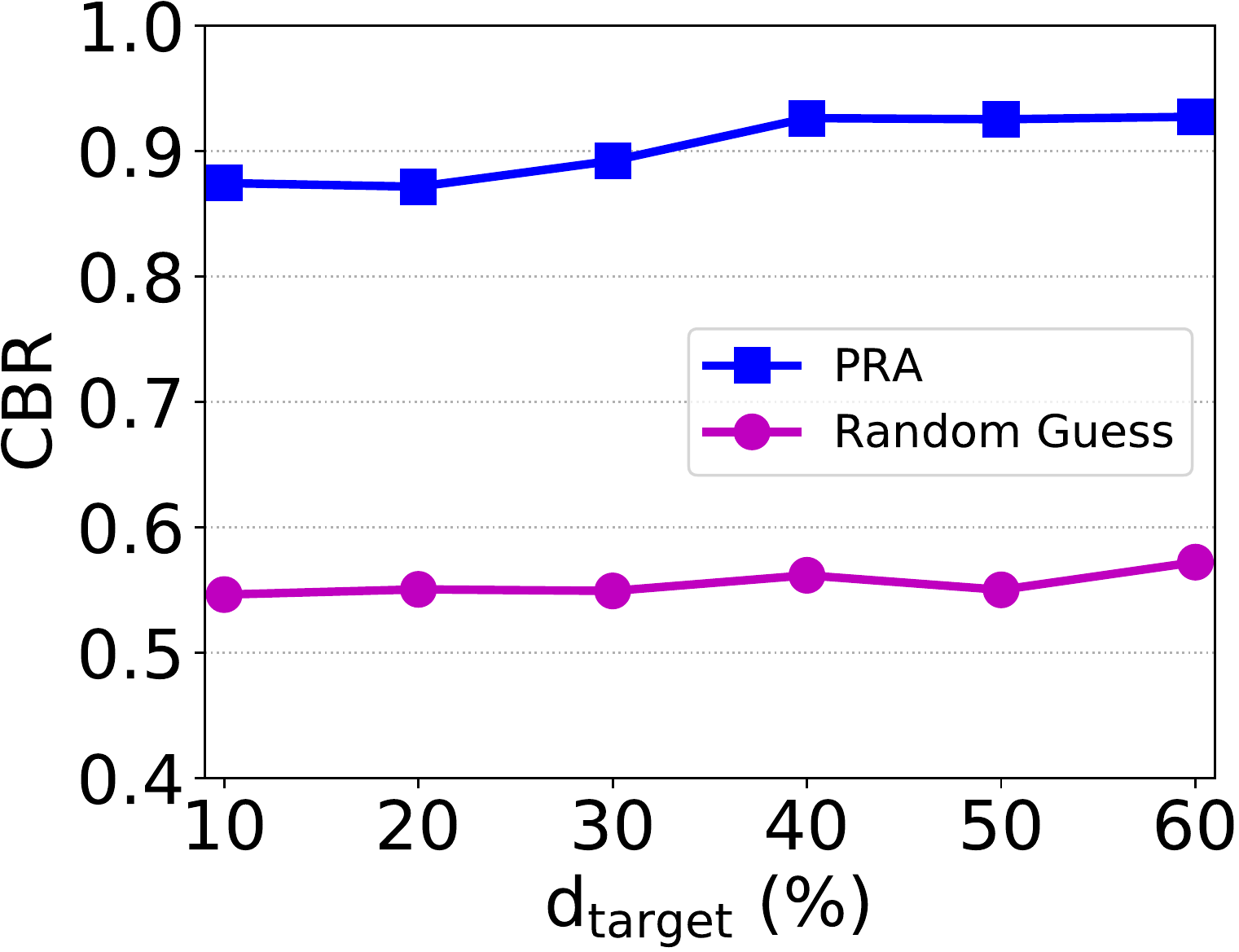}
  \caption{Drive diagnosis}
  \label{subfig:pra-drive}
\end{subfigure}
~
\begin{subfigure}[b]{.49\columnwidth}
  \centering
  \includegraphics[width=0.95\columnwidth]{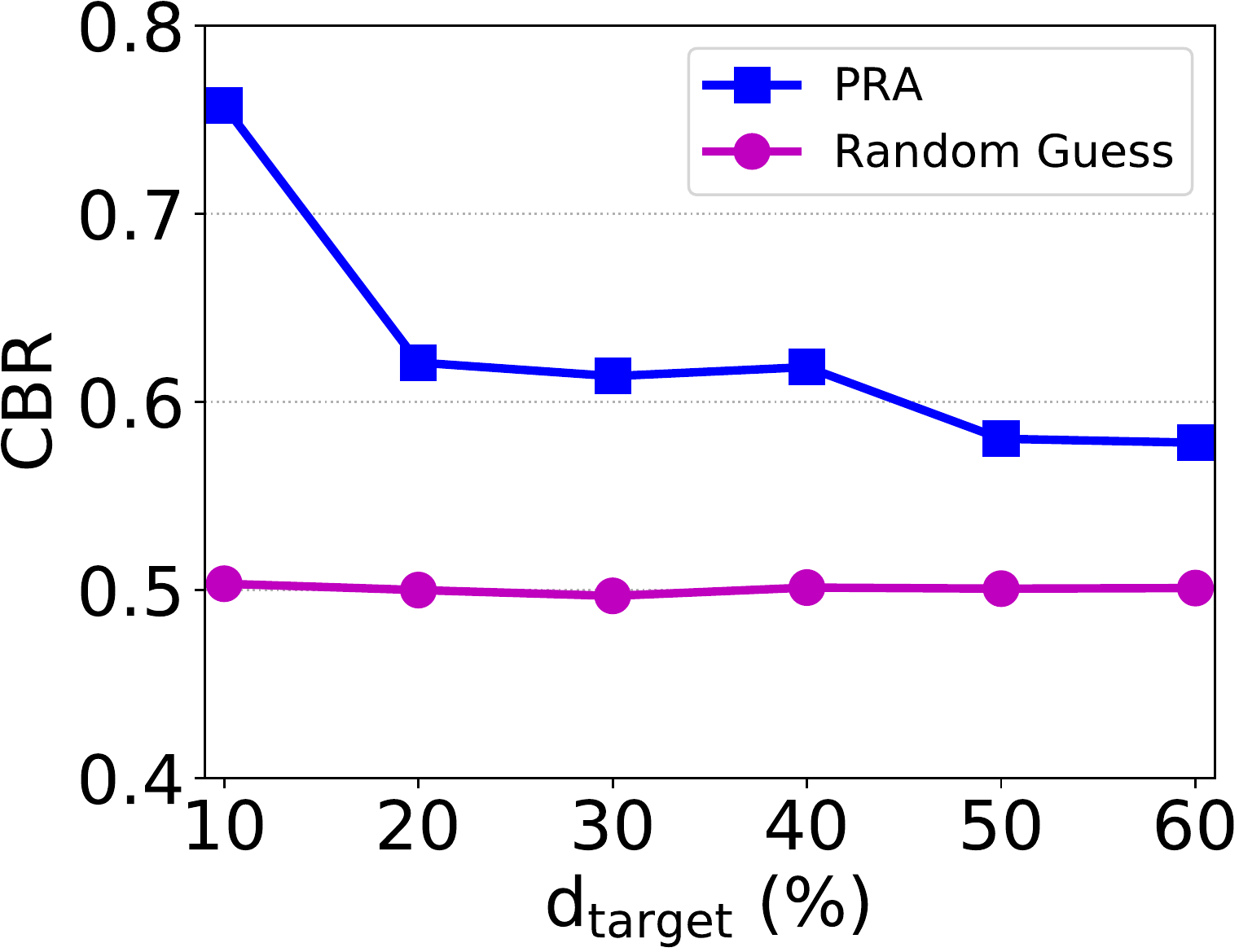}
  \caption{News popularity}
  \label{subfig:pra-news}
\end{subfigure}
\caption{Evaluation of path restriction attack \textit{w.r.t.} CBR}
\vspace{-0.3cm}
\label{fig:pra-performance}
\end{figure*}

\subsection{Experimental Setup} \label{subsec:experiments:setup}

We implement the proposed attack algorithms in Python and conduct experiments on machines that are equipped with Intel (R) Xeon (R) CPU E5-1650 v3 @ 3.50GHz$\times$12 and 32GB RAM, running Ubuntu 16.04 LTS. Specifically, we adopt \textit{PyTorch}\footnote{\url{https://pytorch.org/}}  for training logistic regression (LR) and neural networks (NN) models, and \textit{sklearn}\footnote{\url{https://scikit-learn.org/stable/}} for training decision tree (DT) and random forest (RF) models.

\vspace{1mm}\noindent
\textbf{Datasets.} We evaluate the attack performance on four real-world datasets:
(i) bank marketing dataset \cite{MoroCR14}, which consists of 45211 samples with 20 features and 2 classes;
(ii) credit card dataset \cite{YehL09a}, which consists of 30000 samples with 23 features and 2 classes;
(iii) drive diagnosis dataset \cite{Dua2019}, which consists of 58509 samples with 48 features and 11 classes.
(iv) news popularity dataset \cite{FernandesV015}, which consists of 39797 samples with 59 features and 5 classes.

Besides, we generate two synthetic datasets with the \textit{sklearn} library for evaluating the impact of the number of samples $n$ in the prediction dataset on the performance of generative regression network attack.
The first synthetic dataset includes 100000 samples with 25 features and 10 classes, and the second includes 100000 samples with 50 features and 5 classes.
We summarize the evaluated datasets in Table~\ref{tab-datasets}.
In addition, we normalize the ranges of all feature values in each dataset into $(0, 1)$ before training the models.

\begin{table}[t]
\caption{Statistics of Datasets}
\centering
\begin{tabular}{  c  c c c }
\toprule
Dataset & Sample Num. & Class Num. & Feature Num. \\
\toprule
Bank marketing &	45211&	2&	20\\
Credit card	&30000&	2	&23\\
Drive diagnosis	&58509&	11&	48\\
News popularity	&39797&	5&	59\\
Synthetic dataset 1	& 100000 &	10&	25\\
Synthetic dataset 2	& 100000 &	5&	50\\
\bottomrule
\end{tabular}
\label{tab-datasets}
\end{table}

\vspace{1mm}
\noindent
\textbf{Models.} We generate the vertical FL models using centralized training and give the trained models to the adversary. This is reasonable because we consider the case that no intermediate information is disclosed during the training process, and only the final model is released. We evaluate the attack performance on the four models discussed in this paper. By default, the maximum tree depth of the DT model is set to 5. For the RF model, the number of trees and the maximum tree depth are set to 100 and 3, respectively. Besides, the NN model is composed of an input layer (whose size equals to the number of total features $d$), an output layer (whose size equals to the number of classes $c$), and three hidden layers (with 600, 300, 100 neurons, respectively).

\vspace{1mm}
\noindent
\textbf{Metrics.} We evaluate the attack performance with two metrics.

Since the equality solving attack (ESA) and generative regression network attack (GRNA) are regression tasks, we use the mean square error (MSE) per feature to measure their overall accuracy in reconstructing multiple target features.

Specifically, the MSE per feature is calculated as:
\begin{align}
    \label{eq:mse-per-feature}
    \text{MSE} = \frac{1}{n * d_{\text{target}}} \sum\nolimits_{t=1}^n \sum\nolimits_{i=1}^{d_{\text{target}}} {(\hat{x}_{\text{target},i}^t - x_{\text{target},i}^t)}^2
\end{align}
where $n$ is the number of samples in the prediction dataset, $d_{\text{target}}$ is the number of target features, $\hat{\boldsymbol{x}}_{\text{target}}^t$ and $\boldsymbol{x}_{\text{target}}^t$ are the inferred feature values and the ground-truth of the $t$-th sample, respectively.
For the path restriction attack (PRA), we measure the correct branching rate (CBR). Specifically, we first randomly select a path from all the possible prediction paths computed by PRA, then accordingly measure the fraction of inferred feature values that belong to the correct branches (i.e., comparing to the ground-truth).
For each experiment, we conduct 10 independent trials and report the average result.

\vspace{1mm}
\noindent
\textbf{Baselines.} For ESA and GRNA,
we use two baselines that randomly generate samples from $(0,1)$ according to a Uniform distribution $\boldsymbol{U}(0, 1)$ and a Gaussian distribution $\boldsymbol{N}(0.5, 0.25^2)$. This Gaussian distribution can ensure that at least $95\%$ samples are within $(0, 1)$.
For PRA on the DT model, we adopt a baseline that randomly selects a prediction path and evaluates CBR along that path. Both baselines are called \textit{random guess} in the following presentation.

\subsection{Evaluation of Attacks Based on Individual Prediction} \label{subsec:experiments:individual}

\begin{figure*}[t]
\begin{subfigure}[b]{.49\columnwidth}
  \centering
  \includegraphics[width=0.95\columnwidth]{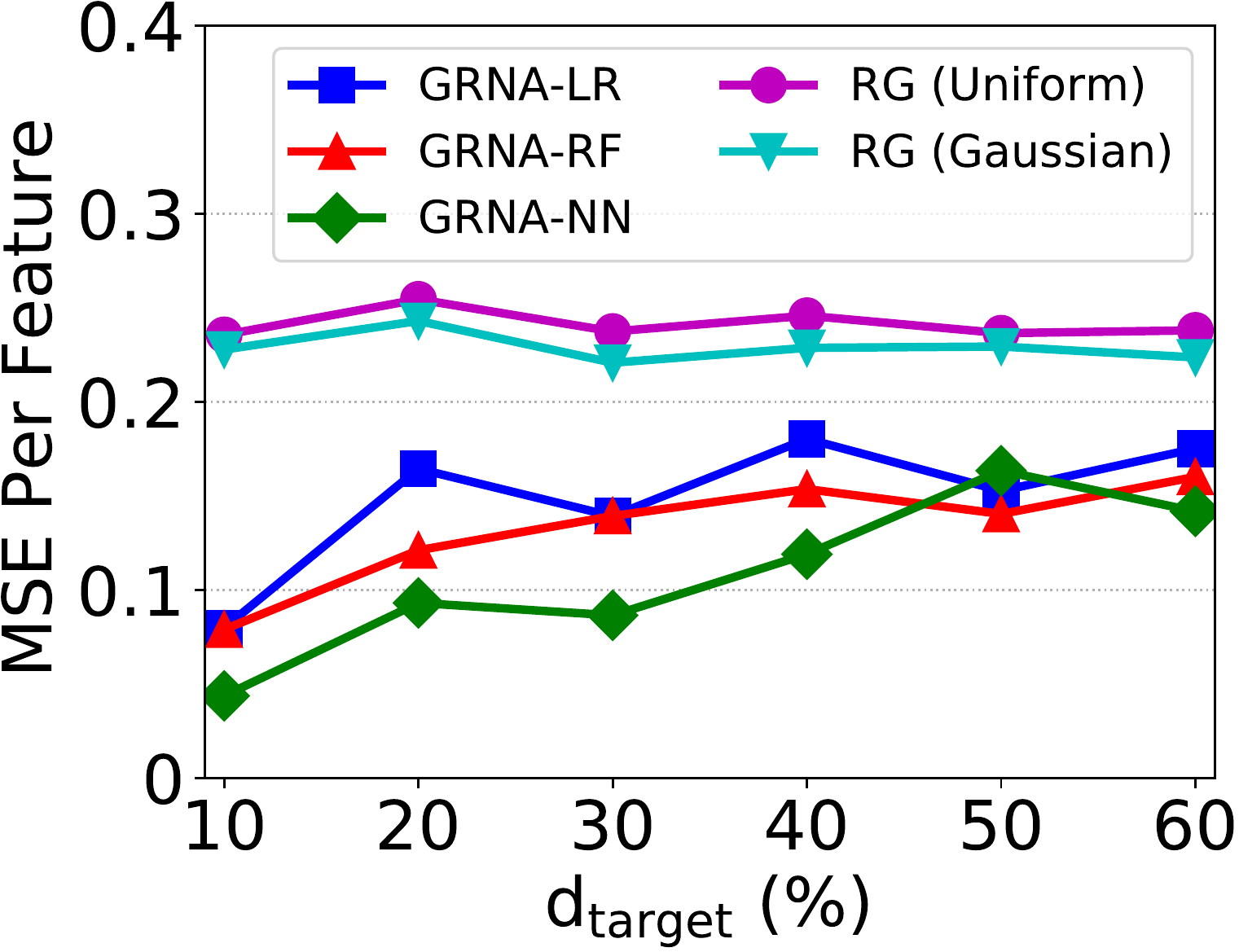}
  \caption{Bank marketing}
  \label{subfig:grna-bank}
\end{subfigure}
~
\begin{subfigure}[b]{.49\columnwidth}
  \centering
  \includegraphics[width=0.95\columnwidth]{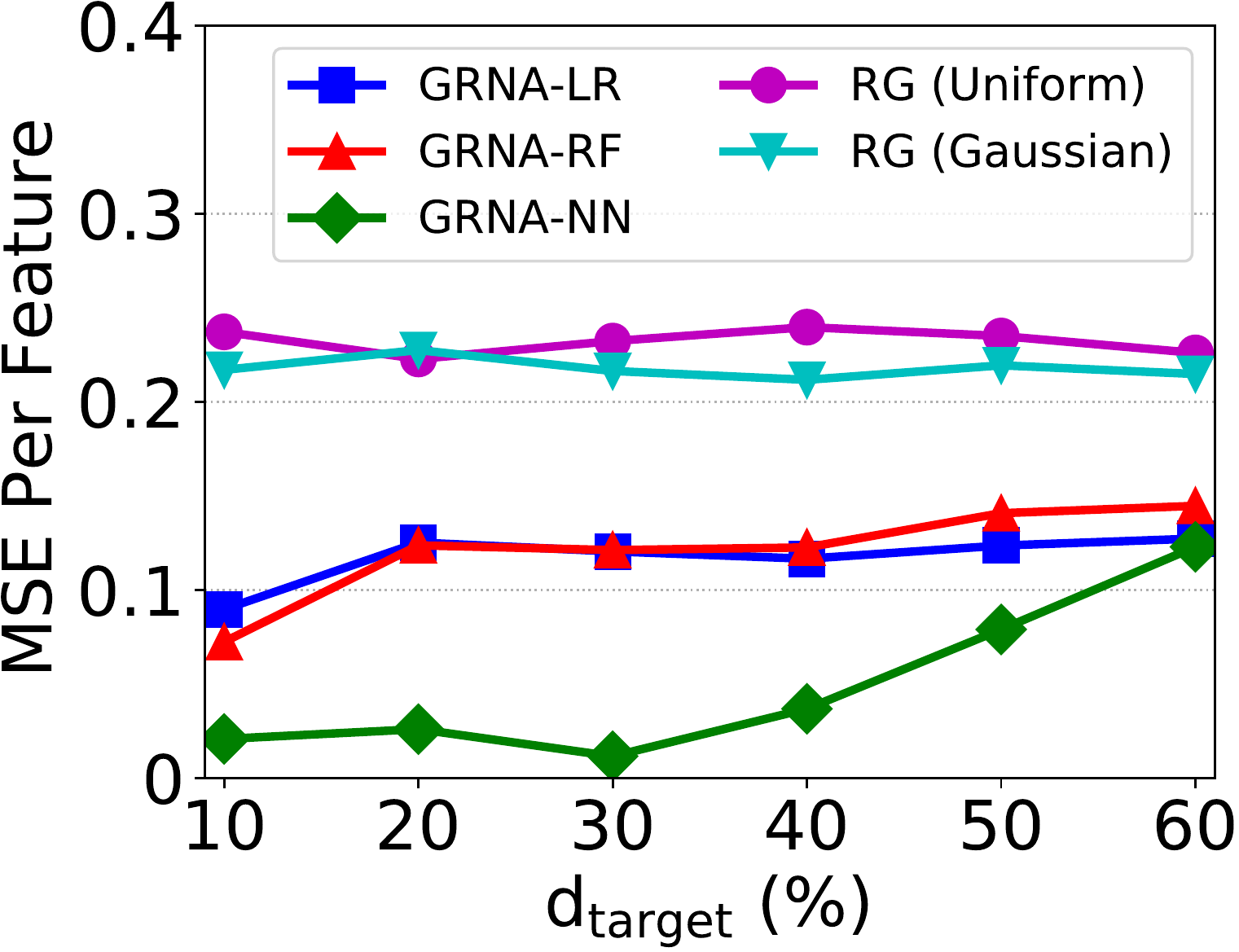}
  \caption{Credit card}
  \label{subfig:grna-credit}
\end{subfigure}
~
\begin{subfigure}[b]{.49\columnwidth}
  \centering
  \includegraphics[width=0.95\columnwidth]{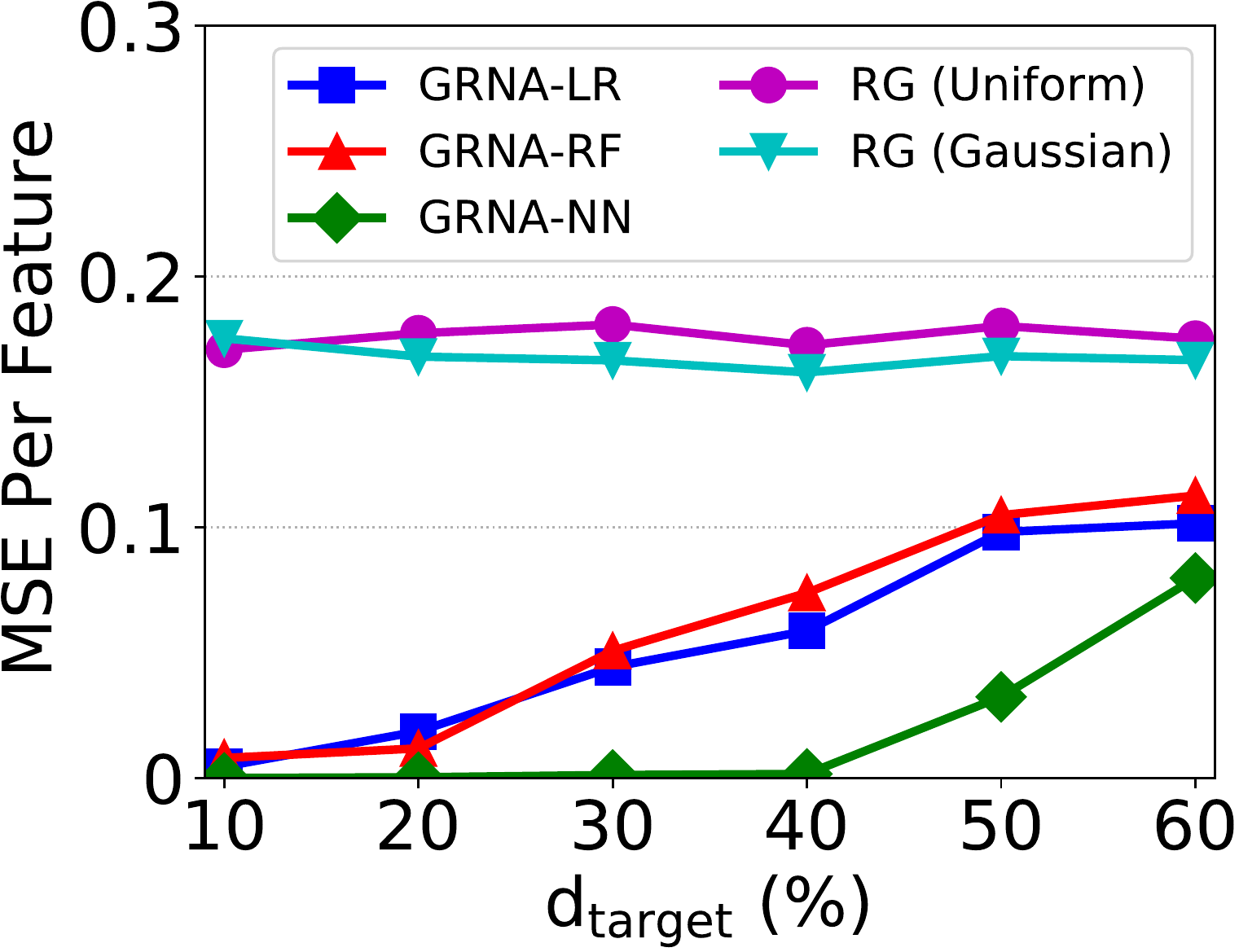}
  \caption{Drive diagnosis}
  \label{subfig:grna-drive}
\end{subfigure}
~
\begin{subfigure}[b]{.49\columnwidth}
  \centering
  \includegraphics[width=0.95\columnwidth]{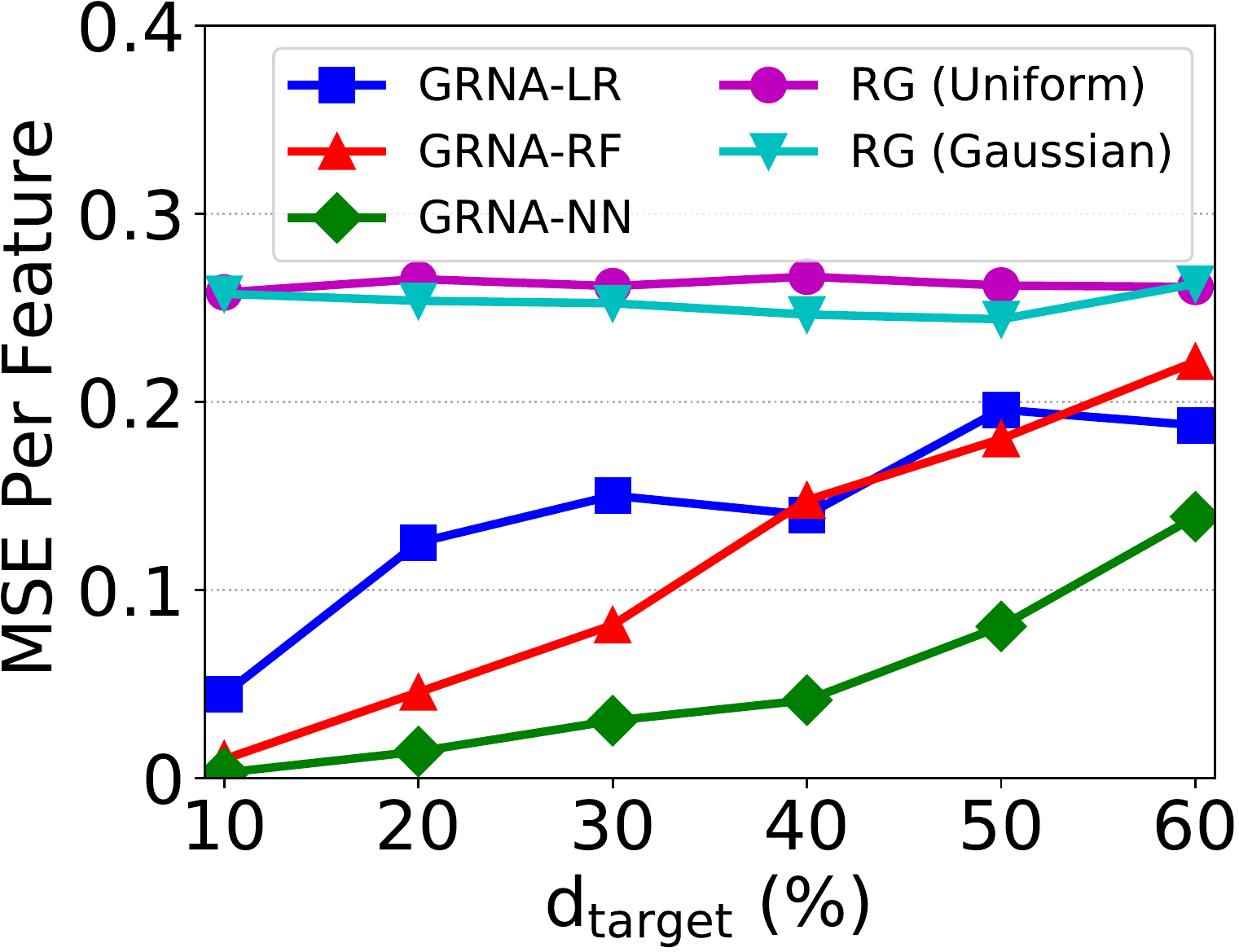}
  \caption{News popularity}
  \label{subfig:grna-news}
\end{subfigure}

\caption{Evaluation of generative regression network attack \textit{w.r.t.} MSE per feature}
\vspace{-0.5cm}
\label{fig:grna-performance}
\end{figure*}

For ESA and PRA based on individual model prediction (see Section~\ref{sec:attacks-individual-prediction}), we evaluate the attack performance \textit{w.r.t.} the number of features $d_{\text{target}}$ owned by the attack target $P_{\text{target}}$. In particular, we vary the fraction of $d_{\text{target}}$ (i.e., over the number of total features $d$) by $\{10\%, 20\%, 30\%, 40\%, 50\%, 60\%\}$.

\vspace{1mm}\noindent
\textbf{Effect of $d_{\text{target}}$ on ESA.} Fig.~\ref{fig:esa-performance} shows the attack performance of ESA \textit{w.r.t.} MSE per feature. For all the datasets, if the threshold condition $d_{\text{target}} \leq c - 1$ is satisfied (denoted by `T' in each sub-figure), the MSE per feature is 0. This is in line with what we have discussed in Section~\ref{subsec:attacks-lr-nn:specfic}. For example, there are 11 classes in the drive diagnosis dataset; thus in Fig.~\ref{subfig:esa-drive}, the unknown feature values can be precisely inferred when $d_{\text{target}} = 10$ (i.e., 20\% in percentage).

Moreover, we observe that even if the threshold condition is not satisfied, ESA can still find a good inference of $\boldsymbol{x}_{\text{target}}$, which is greatly superior to the random guess methods (e.g., in Fig.~\ref{subfig:esa-credit} and \ref{subfig:esa-drive}).
Besides, the MSE per feature increases as the fraction of $d_{\text{target}}$ increases. This is expected since the larger $d_{\text{target}}$ is, the more unknown variables in the equations, making the estimation more biased.

In particular, the MSE increase of the Bank dataset is much larger than those of other datasets, which can be explained by the difference of data distributions of different datasets.
As mentioned in Section~\ref{subsec:attacks-lr-nn:specfic}, the solution $\hat{\boldsymbol{x}}_{\text{target}}$ computed by the pseudo-inverse matrix has the minimum Euclidean norm among all solutions, i.e., $||\hat{\boldsymbol{x}}_{\text{target}}||_2\leq ||\boldsymbol{x}_{\text{target}}||_2$, which indicates
\begin{equation}\label{eq-l2norm}
    \sum^{d_{\text{target}}}_{i=1} \hat{x}_{\text{target}, i}^2\leq \sum^{d_{\text{target}}}_{i=1} x_{\text{target}, i}^2.
\end{equation}
Accordingly, we have
\begin{align}
\text{MSE}&=\frac{1}{d_{\text{target}}}\sum^{d_{\text{target}}}_{i=1} (\hat{x}_{\text{target}, i} - {x}_{\text{target}, i})^2\\
       &=\frac{1}{d_{\text{target}}}\sum^{d_{\text{target}}}_{i=1} (\hat{x}_{\text{target}, i}^2 + {x}_{\text{target}, i}^2 - 2\hat{x}_{\text{target}, i}x_{\text{target}, i})\label{eq-base}\\
       &\leq\frac{1}{d_{\text{target}}}\sum^{d_{\text{target}}}_{i=1} (\hat{x}_{\text{target}, i}^2 + {x}_{\text{target}, i}^2) \label{eq-step1}\\
       &\leq\frac{1}{d_{\text{target}}}\sum^{d_{\text{target}}}_{i=1} 2{x}_{\text{target}, i}^2\label{eq-finalstep},
\end{align}
where Eqn~(\ref{eq-step1}) is derived from the fact that all the elements of $\boldsymbol{x}_{\text{target}}$ are normalized into $(0,1)$.
Further, by substituting Eqn~(\ref{eq-l2norm}) into Eqn~(\ref{eq-step1}), we obtain Eqn~(\ref{eq-finalstep}), which is an upper bound for  $\text{MSE}(\hat{\boldsymbol{x}}_{\text{target}}, \boldsymbol{x}_{\text{target}})$.
%

\begin{table*}[t]
\caption{Results of ablation study based on LR with Bank marketing}
\centering
\begin{tabular}{c c c c c c}
\toprule
Case Index & Input $\boldsymbol{x}_{\text{adv}}$ & Input Noise & Constraint on $\hat{\boldsymbol{x}}_{\text{target}}$ & Generator & MSE \\
\toprule
1&	$\times$&	$\surd$&	$\surd$&	$\surd$&	0.1705  \\
2&	$\surd$&	$\times$&	$\surd$&	$\surd$&	0.1514\\
3&	$\surd$&	$\surd$&	$\times$&	$\surd$&	0.1474\\
4&	$\surd$&	$\surd$&	$\surd$&	$\times$&	0.2862\\
5&	$\surd$&	$\surd$&	$\surd$&	$\surd$&	0.1216\\
6 &	$\times$&	$\times$&	$\times$&	$\times$&	0.2459\\
\bottomrule
\end{tabular}
\label{tab-ablation}
\end{table*}

Upon that, we compute the upper bound for Bank, Credit, Drive and News datasets, which are 0.60, 0.14, 0.45 and 0.34, respectively. In general, the larger the upper bound is, the worse the attack accuracy the adversary could achieve. This explains why the MSE of Bank increases faster than others.
Besides, ESA achieves better results on Drive than on News as the adversary has more equations (i.e., $c-1 = 10$) for Drive than those for News (i.e., $c-1 = 4$). 



\vspace{1mm}\noindent
\textbf{Effect of $d_{\text{target}}$ on PRA.} Fig.~\ref{fig:pra-performance} shows the attack performance of PRA \textit{w.r.t.} CBR. 
Notice that the comparison between PRA and the random guess methods is similar to Fig.~\ref{fig:esa-performance}, where the attack accuracy degrades as the fraction of $d_{\text{target}}$ goes up. In general, more target features would lead to more possible prediction paths. Therefore, the probability for the adversary to select the correct path is reduced.

Note that the CBR of Drive is stable or even slightly increases as $d_{\text{target}}$ increases. There are two reasons.
First, the Drive dataset has 11 classes, which is much more than those in the other datasets (2 in Bank and Credit, 5 in News).
As such, in Drive, each class corresponds to a smaller number of tree paths.
Therefore, given one specific class, there exist only a small number of candidate prediction paths, which leads an improved CBR.
In this case, the adversary would not gain a significant advantage by knowing more features.
%
Second, the decision tree model only selects informative features during training, which means the increase of $d_{\text{target}}$ in Drive does not necessarily increase the number of unknown features in the tree, since some features may never be selected. As a consequence, a larger $d_{\text{target}}$ do not always decrease the CBR of the PRA attack.




\begin{figure*}[t]
\begin{subfigure}[b]{.49\columnwidth}
  \centering
  \includegraphics[width=0.95\columnwidth]{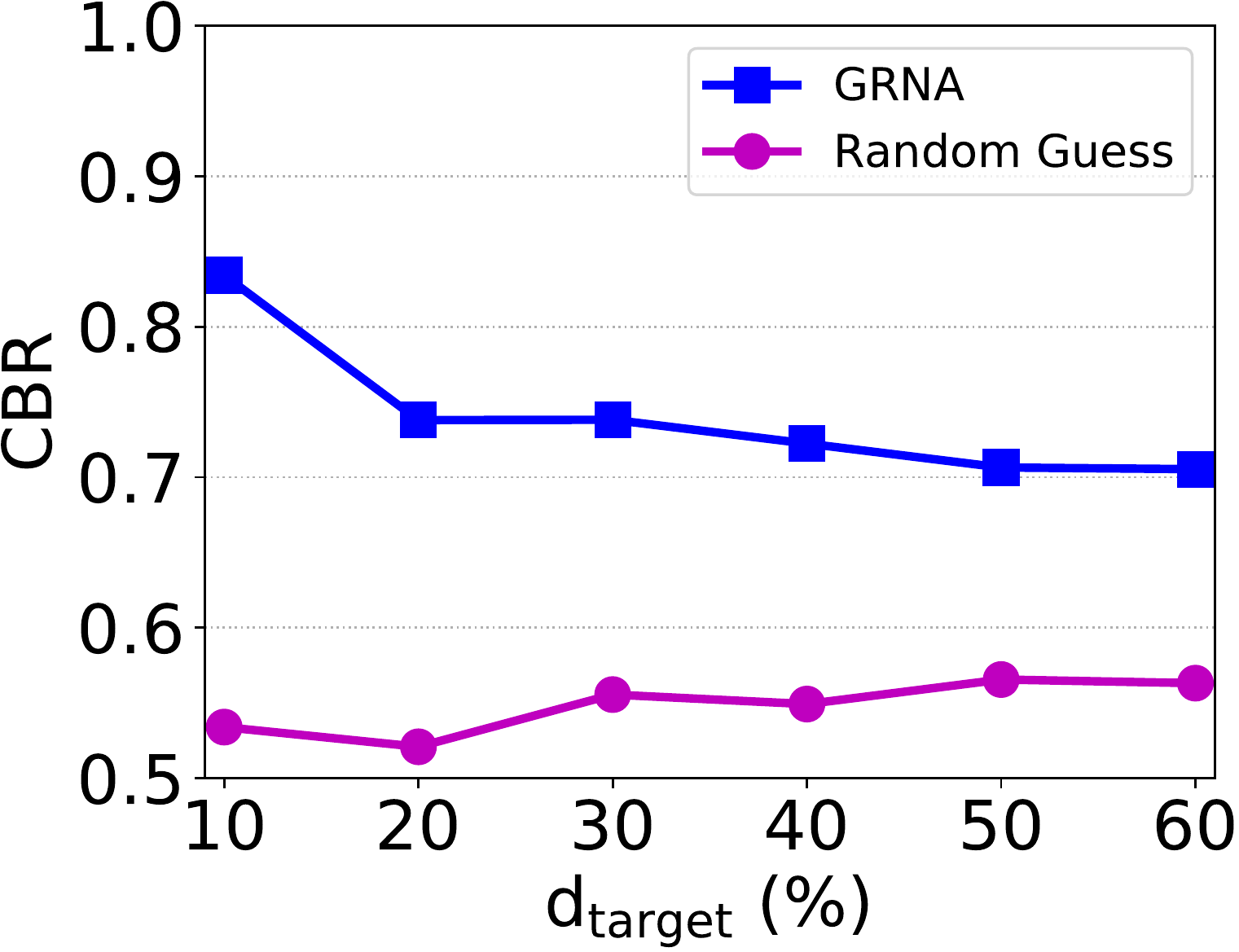}
  \caption{Bank marketing}
  \label{subfig:grna-bank-rf-cbr}
\end{subfigure}
~
\begin{subfigure}[b]{.49\columnwidth}
  \centering
  \includegraphics[width=0.95\columnwidth]{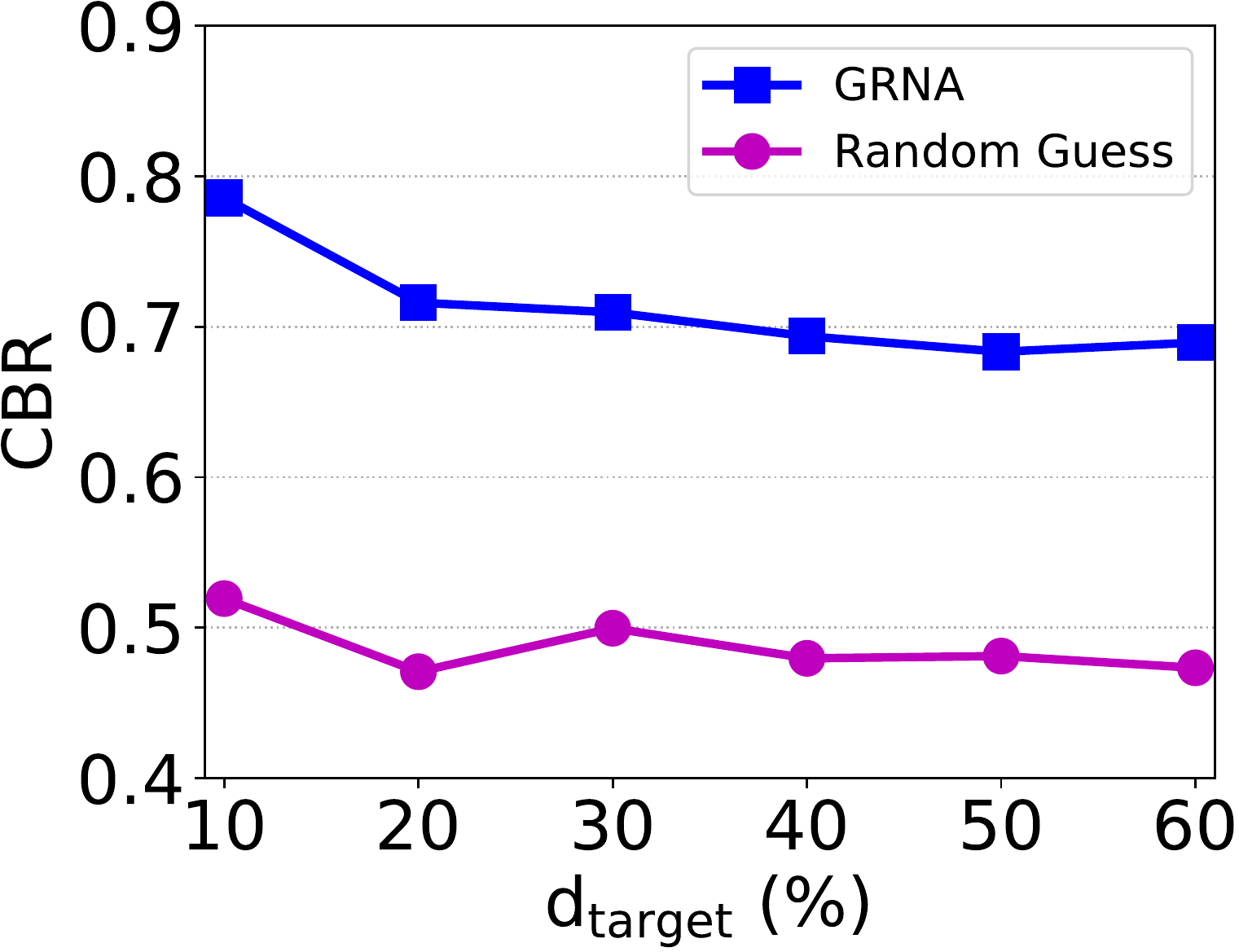}
  \caption{Credit card}
  \label{subfig:grna-credit-rf-cbr}
\end{subfigure}
~
\begin{subfigure}[b]{.49\columnwidth}
  \centering
  \includegraphics[width=0.95\columnwidth]{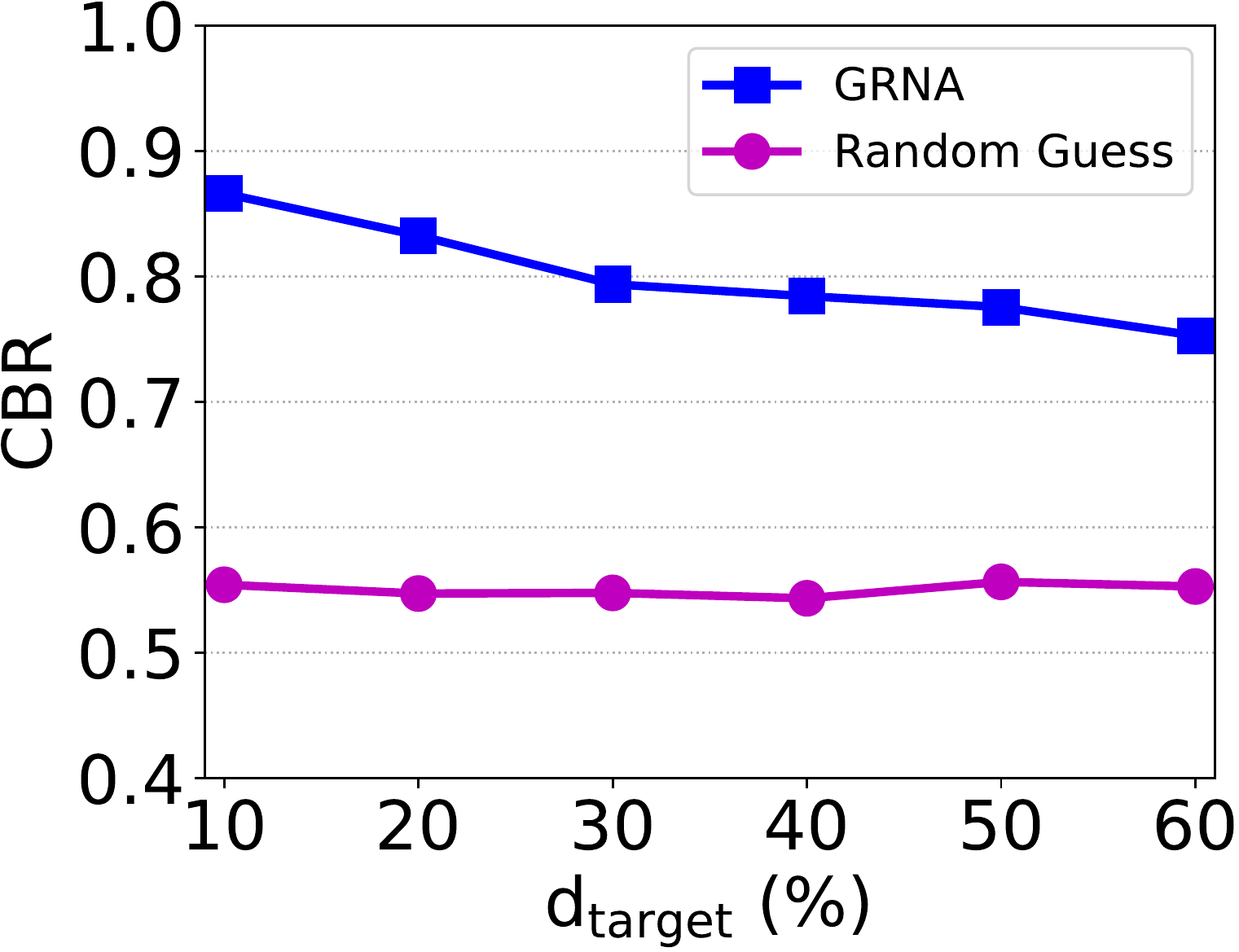}
  \caption{Drive diagnosis}
  \label{subfig:grna-drive-rf-cbr}
\end{subfigure}
~
\begin{subfigure}[b]{.49\columnwidth}
  \centering
  \includegraphics[width=0.95\columnwidth]{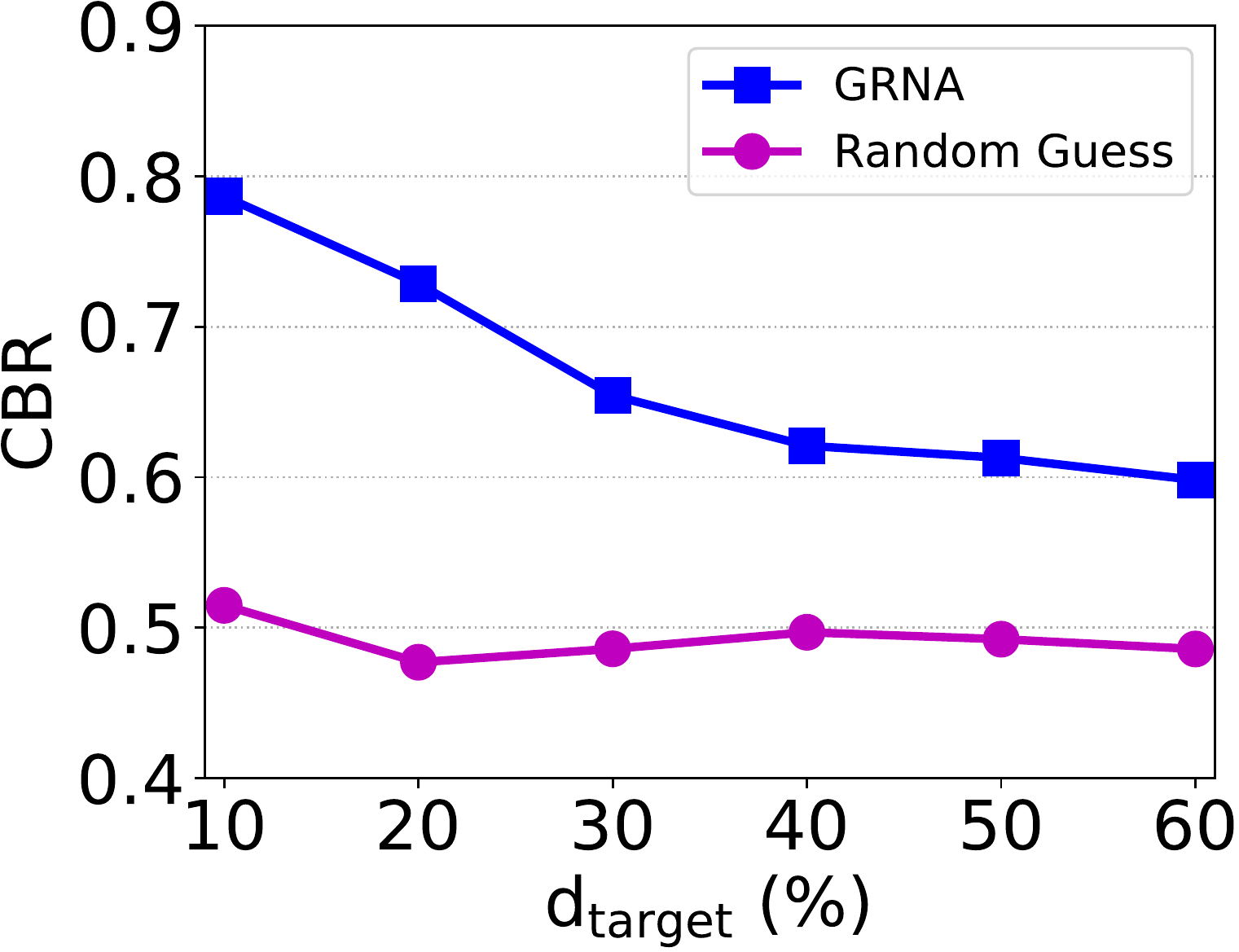}
  \caption{News popularity}
  \label{subfig:grna-news-rf-cbr}
\end{subfigure}

\caption{Evaluation of generative regression network attack on the RF model \textit{w.r.t.} CBR}
\label{fig:grna-rf-performance-cbr}
\end{figure*}

\begin{figure*}[t]
\begin{subfigure}[b]{.48\columnwidth}
  \centering
  \includegraphics[width=0.95\columnwidth]{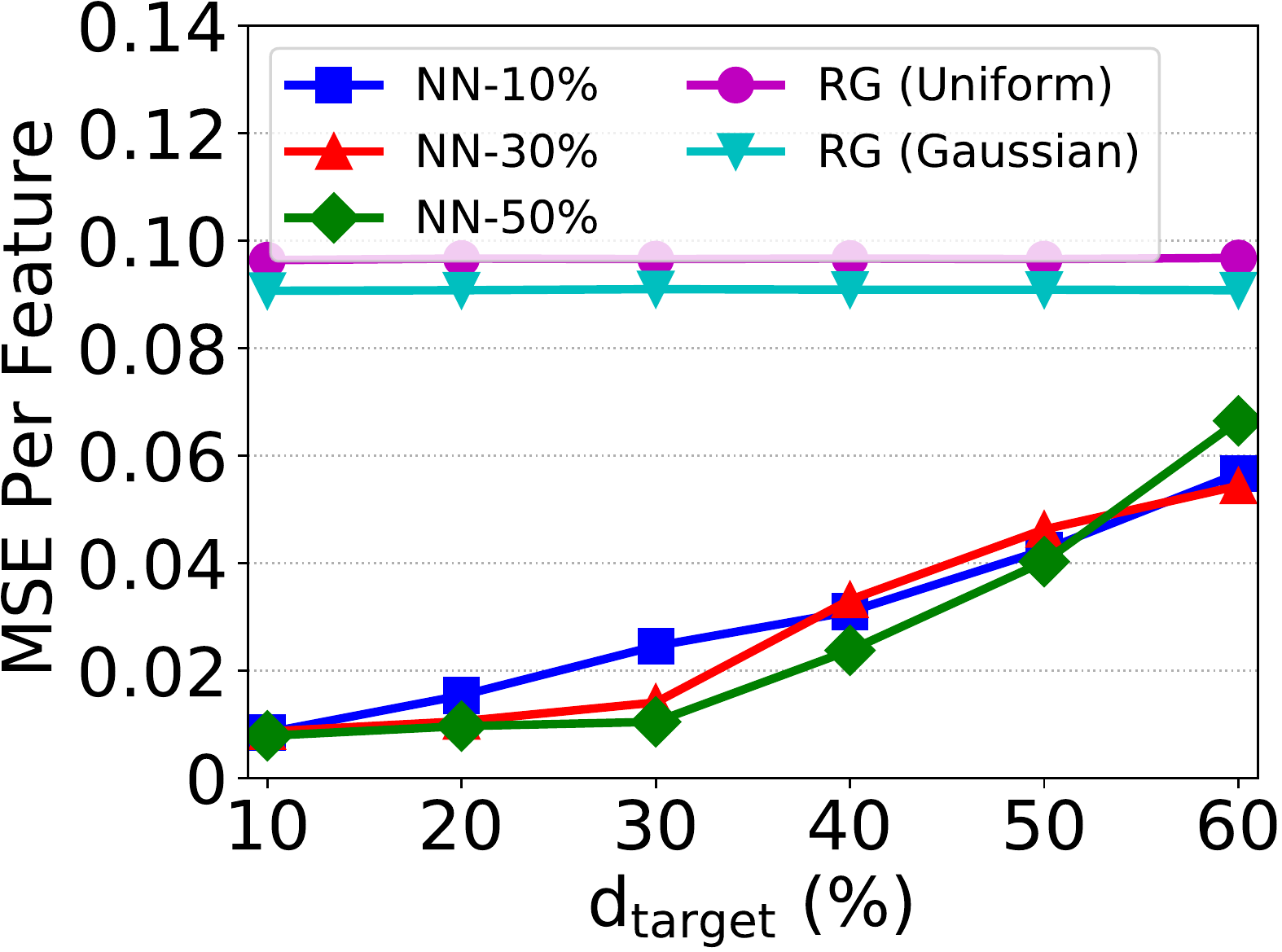}
  \caption{Synthetic dataset 1}
  \label{subfig:synthetic-1}
\end{subfigure}
~
\begin{subfigure}[b]{.48\columnwidth}
  \centering
  \includegraphics[width=0.95\columnwidth]{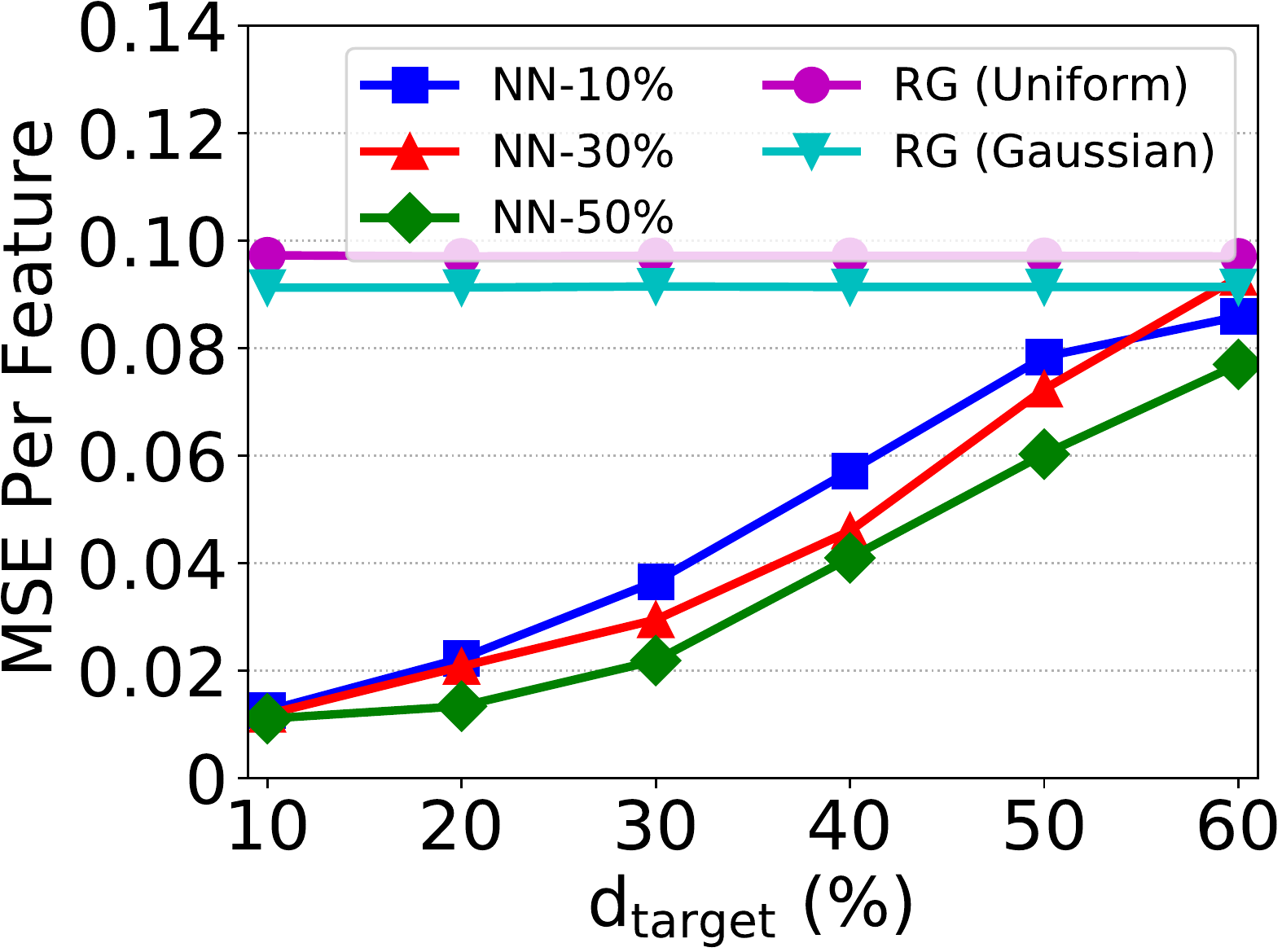}
  \caption{Synthetic dataset 2}
  \label{subfig:synthetic-2}
\end{subfigure}
~
\begin{subfigure}[b]{.48\columnwidth}
  \centering
  \includegraphics[width=0.95\columnwidth]{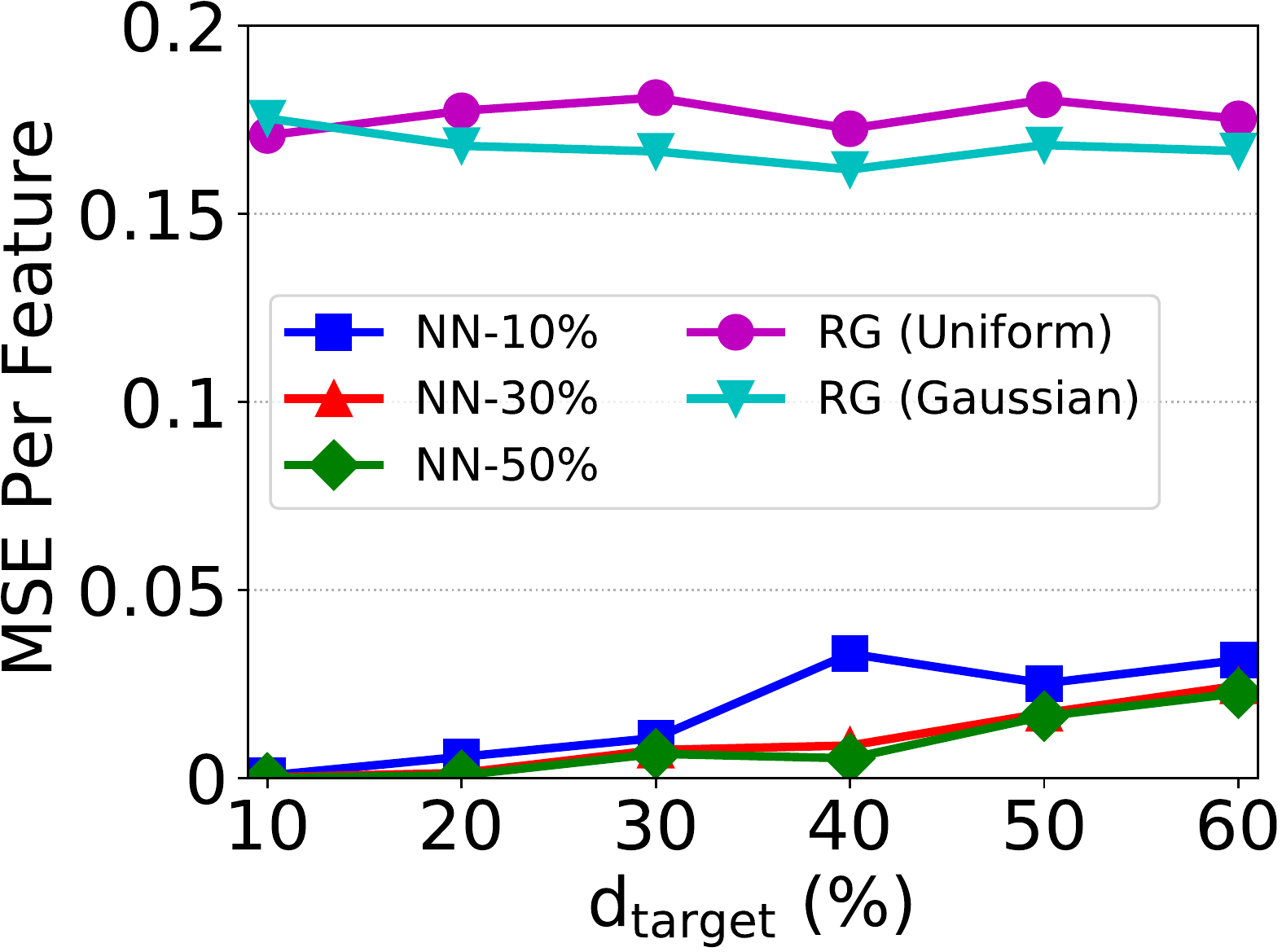}
  \caption{Drive diagnosis}
  \label{subfig:drive-size}
\end{subfigure}
~
\begin{subfigure}[b]{.48\columnwidth}
  \centering
  \includegraphics[width=0.95\columnwidth]{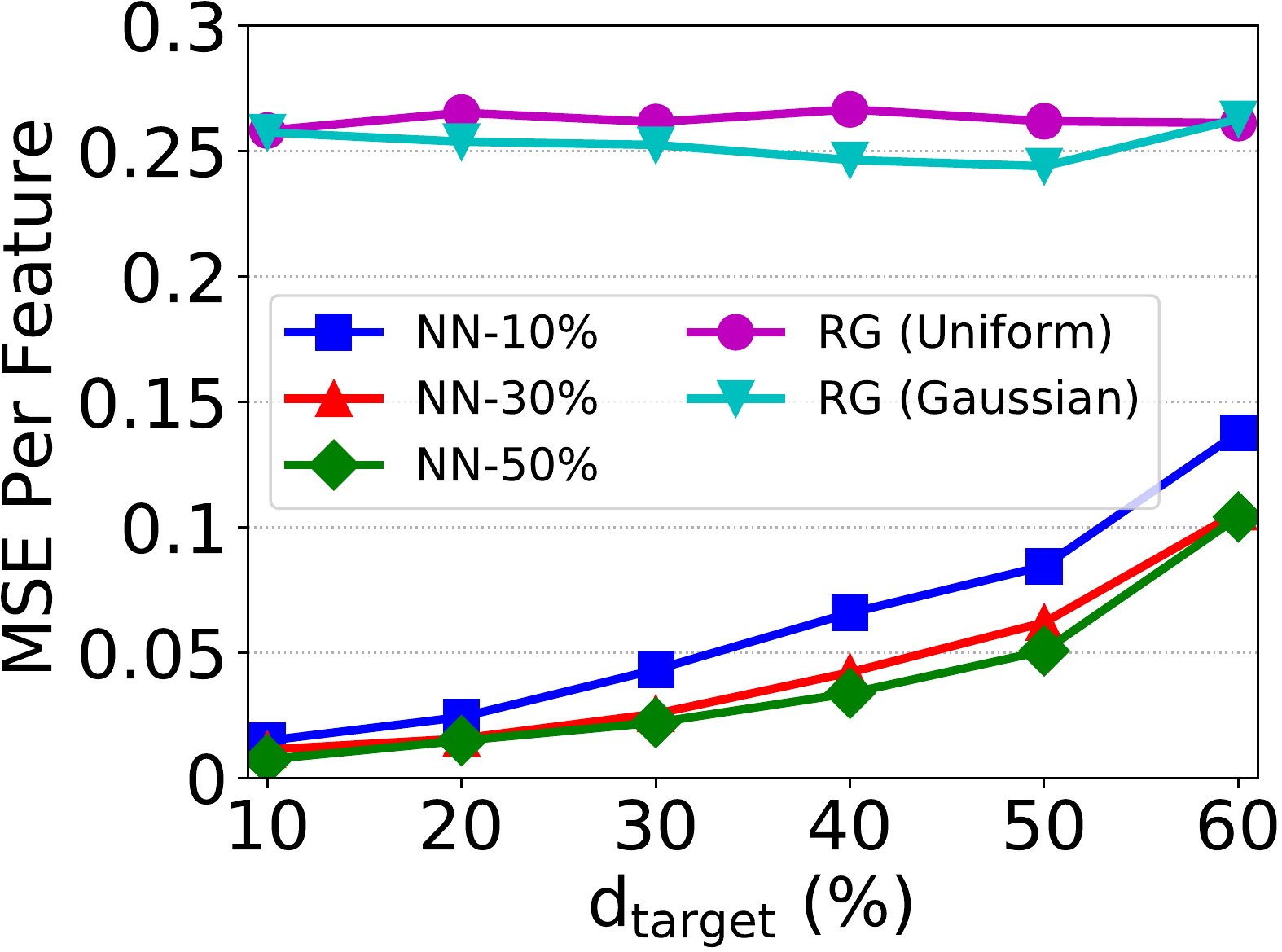}
  \caption{News popularity}
  \label{subfig:news-size}
\end{subfigure}
\caption{Effect of number of predictions \textit{w.r.t.} MSE per feature}
\label{fig:grn-sample-size}
\end{figure*}

\subsection{Evaluation of Attacks Based on Multiple Predictions}
\label{subsec:experiments:multiple}


The generator model of GRNA is a multilayer perceptron with an input layer with size $d$, an output layer with size $d_{\text{target}}$, and three hidden layers (with 600, 200, 100 neurons, respectively). Besides, we employ Layer Normalization~\cite{ba2016layer} after each hidden layer to stabilize the hidden states. For the model that imitates the RF model (see Section~\ref{subsec:attack-rf}), inspired by~\cite{biau2019neural}, we apply another multilayer perceptron with two hidden layers (with 2000 and 200 neurons, respectively).

\vspace{1mm}\noindent
\textbf{Ablation Study on GRNA.}
We first conduct ablation studies on GRN to test the impacts of its different components on the attack performance. The training dataset is Bank marketing, target model is LR, and $40\%$ features of Bank is randomly selected as the $\boldsymbol{x}_{\text{target}}$. We show the experimental results in Table~\ref{tab-ablation}.
In general, we test the following branches: the inputs of the generator are exclusively noises (case 1); the input of the generator is exclusively $\boldsymbol{x}_{\text{adv}}$ (case 2); no converge constraints are added to the generator output $\hat{\boldsymbol{x}}_{\text{target}}$ (case 3); the generator is removed, i.e., a naive regression model which infers $\boldsymbol{x}_{\text{target}}$ based solely on the federated model $f$ and the model output $v$  (case 4). In addition, we show the results of the original GRN (case 5) and random guess (case 6) for comparison.

From Table~\ref{tab-ablation}, we observe that the naive regression model performs worse than random guess (case 4). The reason is that without constraints of $\boldsymbol{x}_{\text{adv}}$, the inferred values of $\hat{\boldsymbol{x}}_{\text{target}}$ regressed from the federated model tend to diverge. We also see that the performance of the generator without $\boldsymbol{x}_{\text{adv}}$ as input degrades most (case 1), which is expected since the effectiveness of GRN mainly relies on the correlations between $\boldsymbol{x}_{\text{adv}}$ and $\boldsymbol{x}_{\text{target}}$. As mentioned in Section~\ref{subsec:attacks-grn}, we penalize the variance of  $\hat{\boldsymbol{x}}_{\text{target}}$ when it it too large to prevent GRN from generating meaningless samples. Case 3 shows that the constraint on $\hat{\boldsymbol{x}}_{\text{target}}$ can reduce the inference error by $17\%$.
In addition, we observe that the random vector, which is concatenated with $\boldsymbol{x}_{\text{adv}}$ as the input of GRN, can further reduce the reconstruction error by $20\%$ (case 2).

\vspace{1mm}\noindent
\textbf{Effect of $d_{\text{target}}$ on GRNA.}
Fig.~\ref{fig:grna-performance} shows the attack performance of GRNA for three models (i.e., LR, RF, and NN) on four real-world datasets \textit{w.r.t.} MSE per feature. Similarly, we vary the fraction of $d_{\text{target}}$.
The trends of the three models on all the datasets are similar, i.e., the MSE per feature all goes up as the fraction of $d_{\text{target}}$ increases. This is because GRNA relies on the feature correlations between $\boldsymbol{x}_{\text{adv}}$ and $\boldsymbol{x}_{\text{target}}$ to infer the unknown feature values. The learned correlations would become weaker if the fraction of $d_{\text{target}}$ is larger, leading to relatively worse attack performance. However, even when the fraction of $d_{\text{target}}$ is 60\%, the GRNA still achieves a much better inference than the random guess methods, demonstrating its effectiveness.
In addition, GRNA with the NN model performs better than that with the LR and RF models. The reason is that the NN model has more complicated decision boundaries than the other two models, thus greatly limiting the possible distributions of ${\boldsymbol{x}}_{\text{target}}$ given the same $\boldsymbol{x}_{\text{adv}}$ and $\boldsymbol{v}$. Meanwhile, with more parameters, the NN model itself can capture more important information about the feature correlations than others, resulting in better attack performance.

As the NN model imitating the RF model only approximates the feature thresholds on the internal tree nodes, we also apply the CBR metric for evaluating GRNA on the RF model. Fig.~\ref{fig:grna-rf-performance-cbr} illustrates the performance for varying the fraction of $d_{\text{target}}$.
Results demonstrate that GRNA recovers many more branches in the trees than the random guess methods. For example, if the fraction of $d_{\text{target}}$ is 10\%, GRNA correctly infers more than 80\% of the tree branches \textit{w.r.t.} the bank marketing and drive diagnosis datasets given the generated feature values.

\vspace{1mm}\noindent
\textbf{Effect of $n$ in the prediction dataset on GRNA.} As the generator model is trained on multiple predictions,
we investigate the effect of the number of predictions $n$ using two synthetic datasets and two real-world datasets.
For a dataset $\mathcal{D}$, we first use half of the dataset for model training and testing, then randomly select $n = \{10\%, 30\%, 50\%\}\times |\mathcal{D}|$ samples from the remaining part as the prediction dataset to train the generator using GRNA.
Fig.~\ref{fig:grn-sample-size} shows the attack performance \textit{w.r.t.} MSE per feature. Overall speaking, the trends on the four datasets demonstrate that the more samples in the prediction dataset, the less MSE per feature the adversary can obtain.
In other words, the adversary could accumulate more prediction outputs in the long term to improve his attack accuracy.

\begin{figure*}
\centering
    \begin{subfigure}[b]{0.4\textwidth}
            \includegraphics[width=\columnwidth]{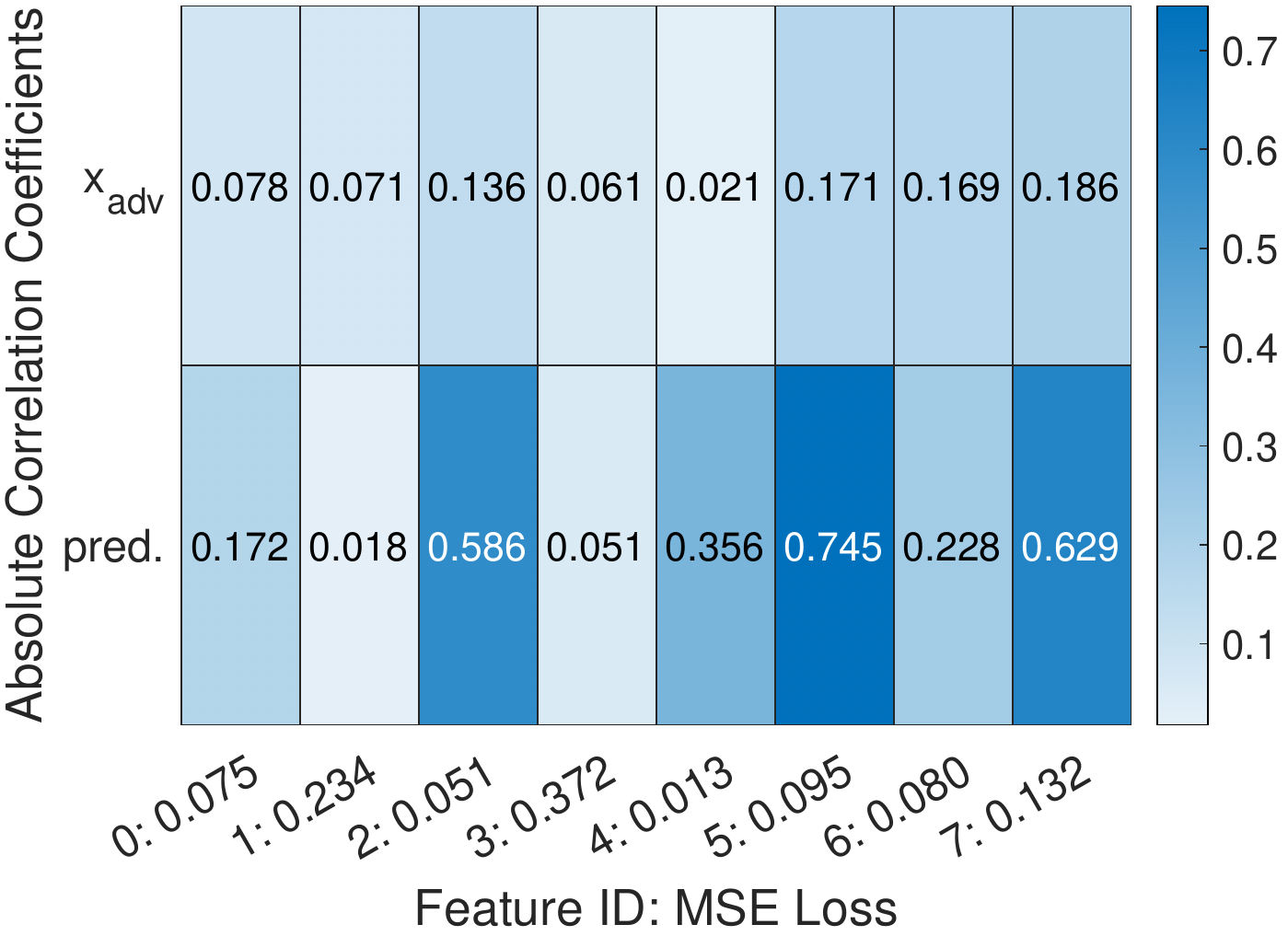}
            \caption{Bank marketing (LR model)}
            \label{fig-bank-corr}
    \end{subfigure}%
    \hspace{2mm}
    \begin{subfigure}[b]{0.4\textwidth}
            \centering
            \includegraphics[width=\columnwidth]{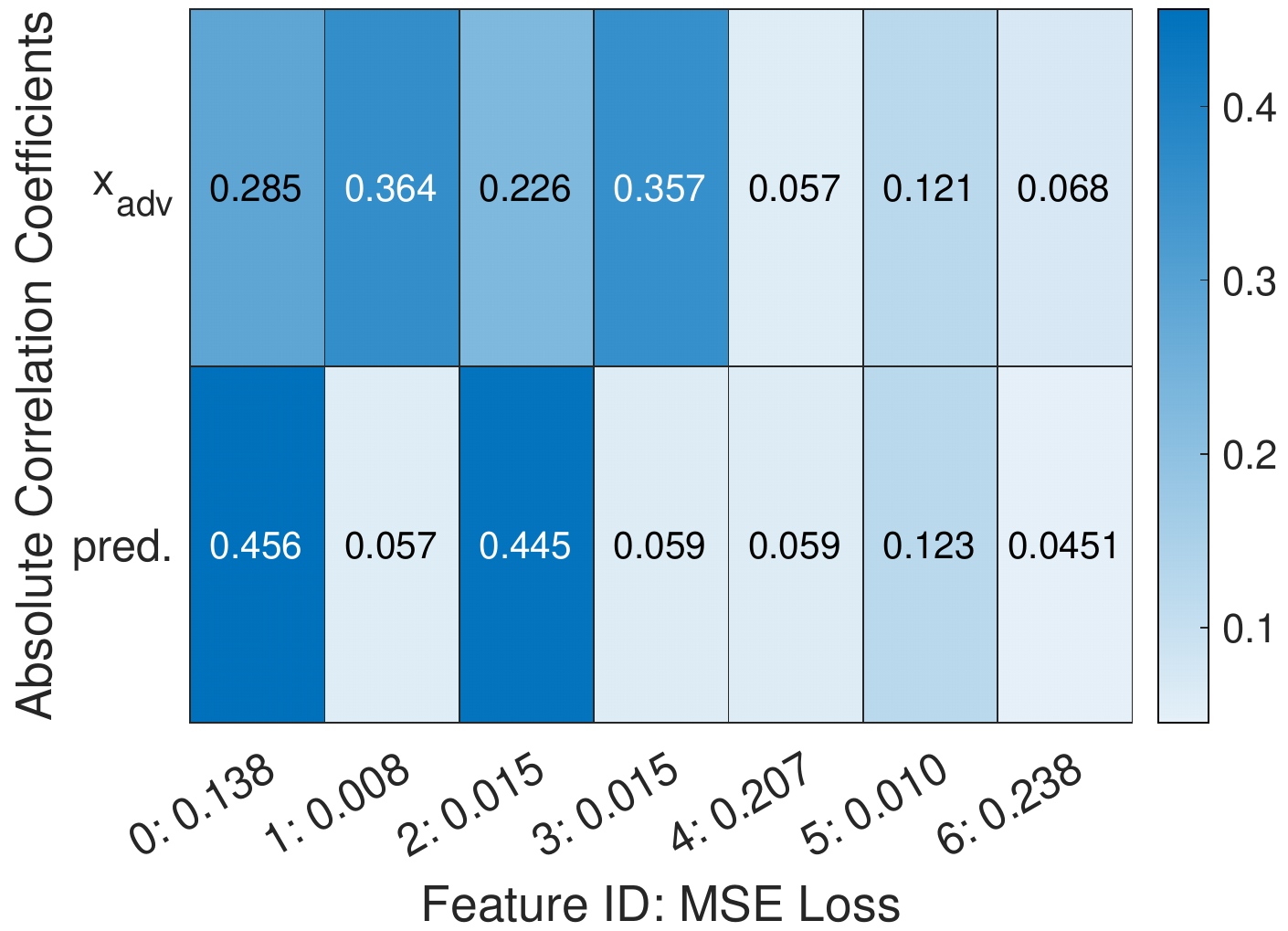}
            \caption{Credit card (RF model)}
            \label{fig-credit-corr}
    \end{subfigure}
    \caption{Effect of data correlations between each feature in $\boldsymbol{x}_{\text{target}}$ with $\boldsymbol{x}_{\text{adv}}$ and $\boldsymbol{v}$}
    \label{fig-corre}
\end{figure*}

\begin{figure*}
\centering
    \begin{subfigure}[b]{0.3\textwidth}
            \includegraphics[width=\columnwidth]{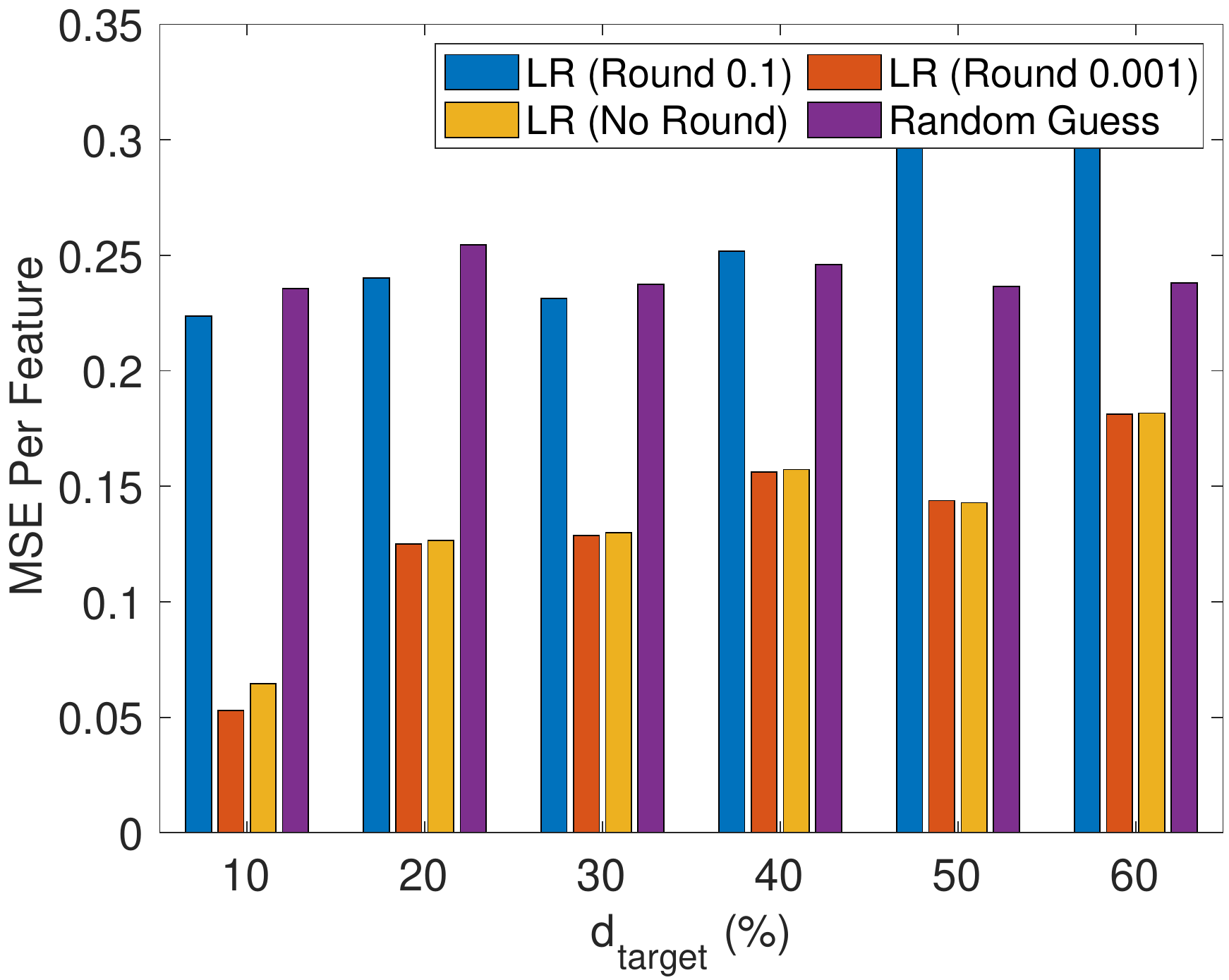}
            \caption{ESA on bank marketing}
            \label{subfig:bank-conf-esa}
    \end{subfigure}%
    ~
    \hspace{2mm}
    \begin{subfigure}[b]{0.3\textwidth}
            \centering
            \includegraphics[width=\columnwidth]{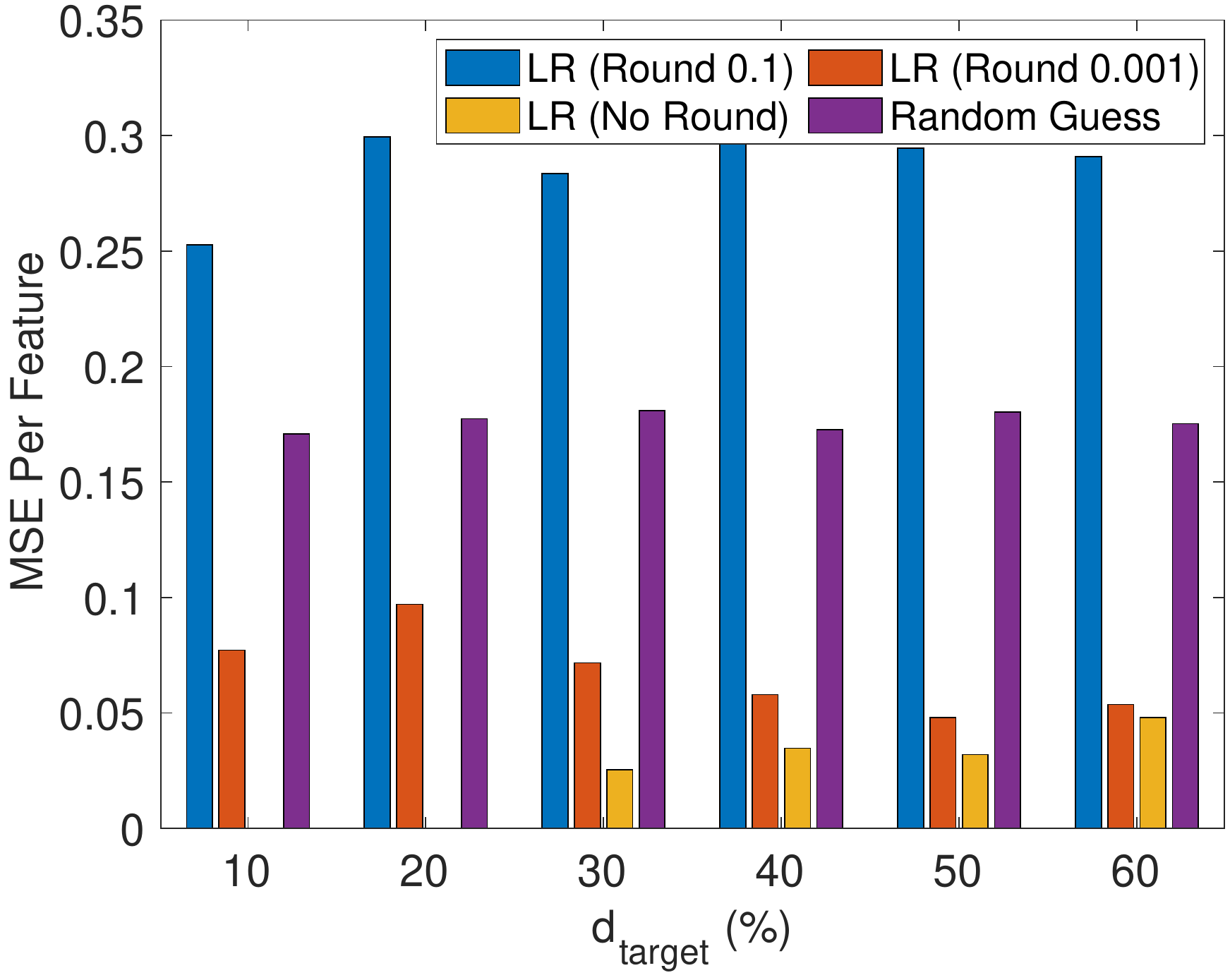}
            \caption{ESA on drive diagnosis}
            \label{subfig:drive-conf-esa}
    \end{subfigure}
    ~
    \hspace{2mm}
    \begin{subfigure}[b]{0.3\textwidth}
            \includegraphics[width=\columnwidth]{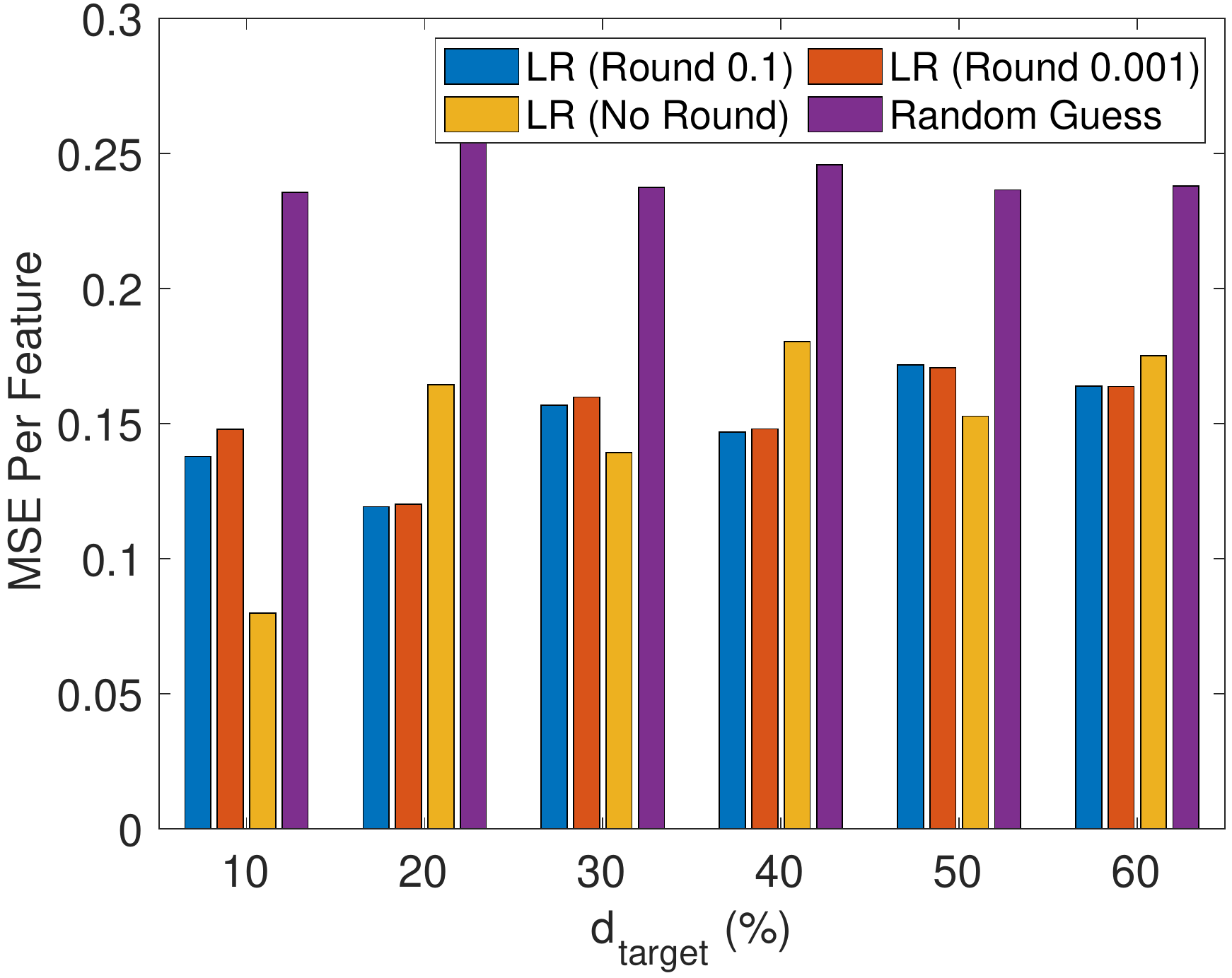}
            \caption{GRNA on bank marketing}
            \label{subfig:bank-conf-grn}
    \end{subfigure}%

\vspace{2mm}
\centering
    \begin{subfigure}[b]{0.3\textwidth}
            \includegraphics[width=\columnwidth]{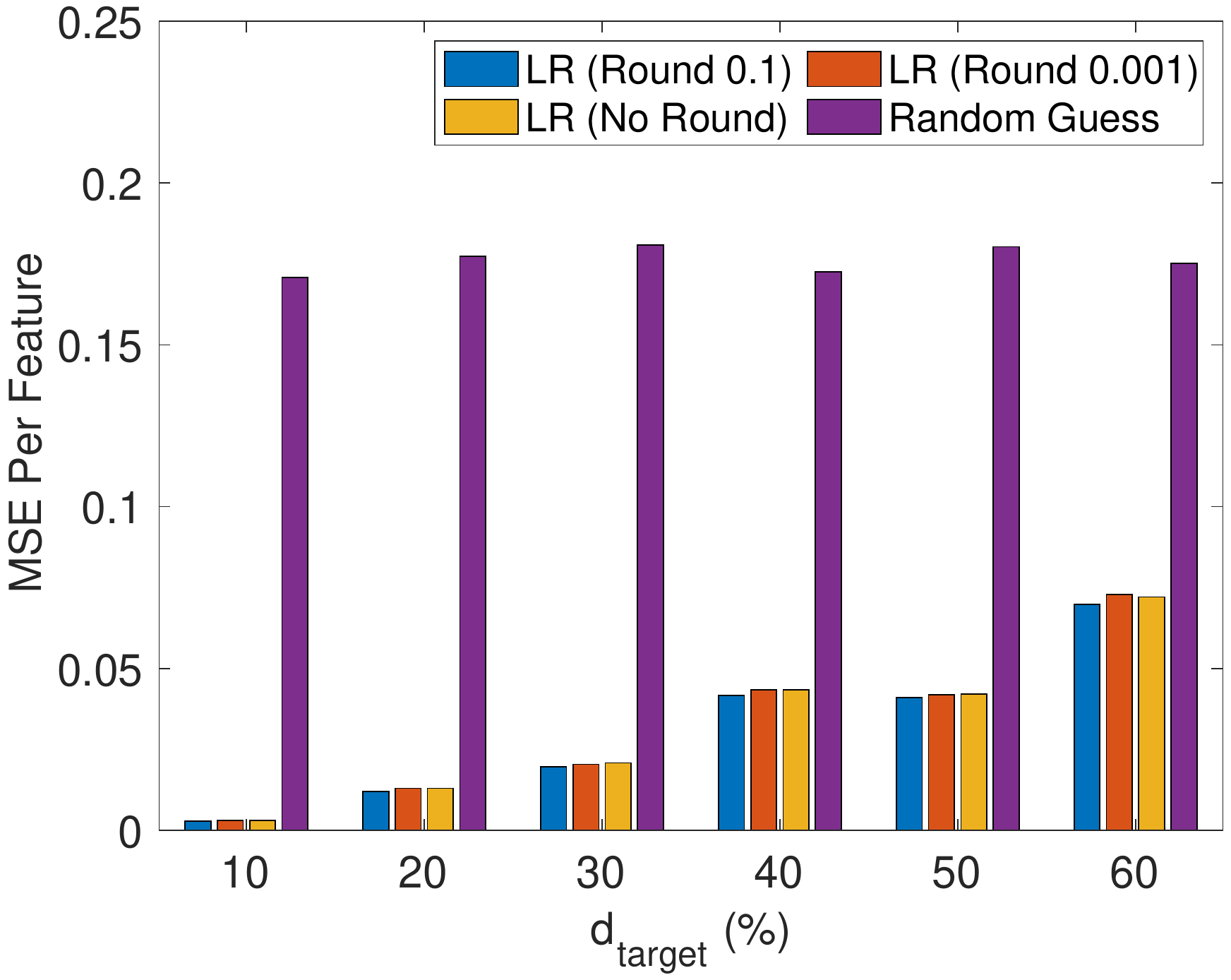}
            \caption{GRNA on drive diagnosis}
            \label{subfig:drive-conf-grn}
    \end{subfigure}%
    ~
    \hspace{2mm}
    \begin{subfigure}[b]{0.3\textwidth}
            \includegraphics[width=\columnwidth]{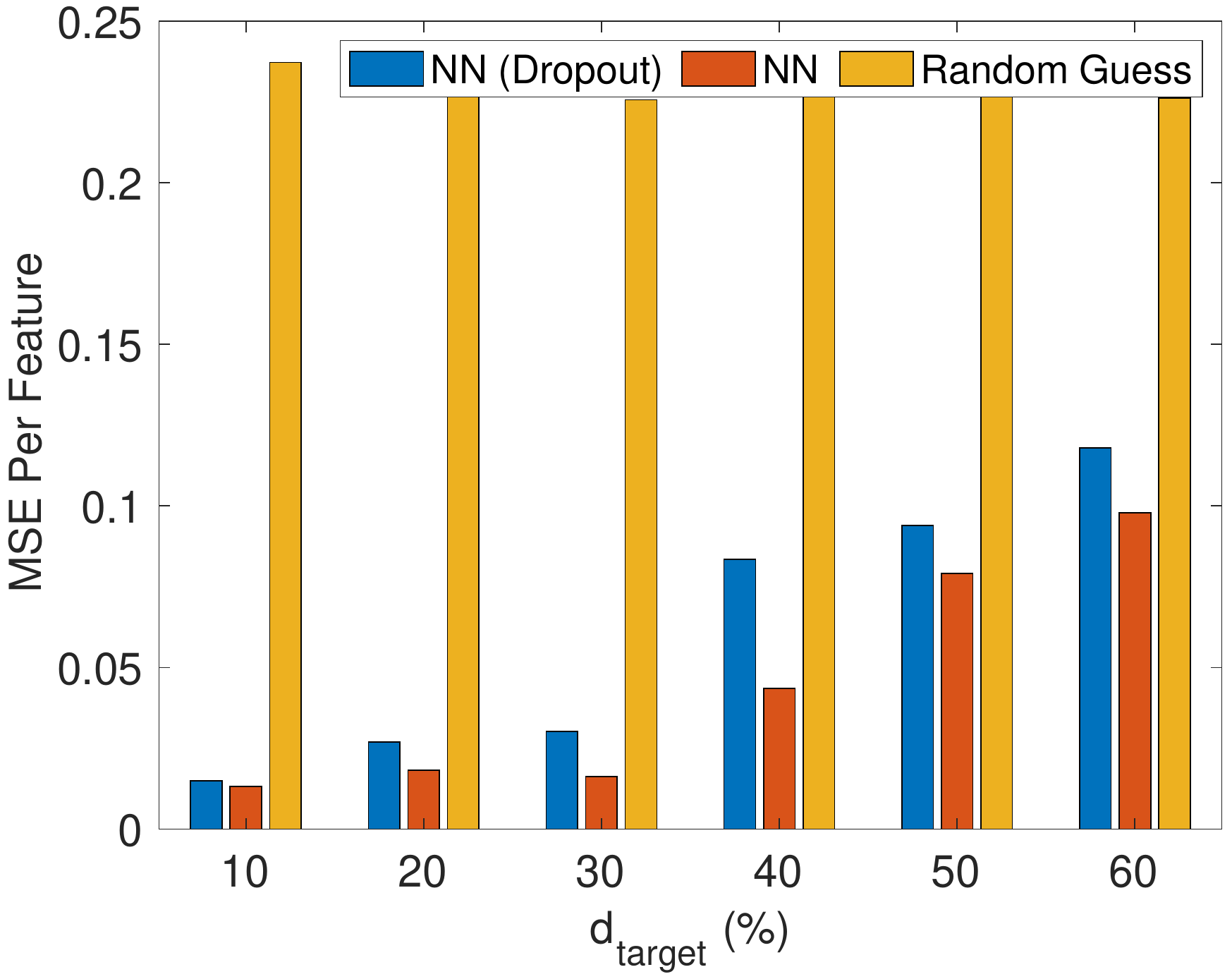}
            \caption{GRNA on credit card}
            \label{subfig:credit-dropout-grn}
    \end{subfigure}%
    ~
    \hspace{2mm}
    \begin{subfigure}[b]{0.3\textwidth}
            \centering
            \includegraphics[width=\columnwidth]{figures/fig-dropout-credit-grn.pdf}
            \caption{GRNA on news popularity}
            \label{subfig:news-dropout-grn}
    \end{subfigure}
    \caption{Evaluation of countermeasures: (a)-(b) and (c)-(d) illustrate the effects of the rounding strategy in ESA and GRNA \textit{w.r.t.} the LR model, respectively; (e)-(f) illustrate the effect of the dropout strategy in GRNA \textit{w.r.t.} the NN model.}
    \label{fig-countermeasures}
    \vspace{-0.5cm}
\end{figure*}

\vspace{1mm}\noindent
\textbf{Effect of data correlations on GRNA.}
Notice that the performances of the LR and RF models are lower than that of the NN model (see Fig.~\ref{fig:grna-performance}).
The reason is that a small part of the inferred feature values is far from the ground-truth, leading to a relatively low overall attack performance.
To further explore this phenomenon, we analyze the impact of data correlations between each target feature in $\boldsymbol{x}_{\text{target}}$ and the adversary's features $\boldsymbol{x}_{\text{adv}}$ as well as the prediction output $\boldsymbol{v}$.
%
Specifically, the data correlation is defined as
\begin{align}
  \label{eq-adv}
  \textit{corr}(\boldsymbol{x}_{\text{adv}}, \boldsymbol{x}_{\text{target}, i}) & = \frac{1}{d_{\text{adv}}} \sum\limits^{d_{\text{adv}}}_{j=1} abs(r(\boldsymbol{x}_{\text{adv}, j}, \boldsymbol{x}_{\text{target}, i})), \\
   \label{eq-yhat}
   \textit{corr}(\boldsymbol{v}, \boldsymbol{x}_{\text{target}, i}) & = \frac{1}{c}\sum\limits^{c}_{j=1} abs(r(v_j, \boldsymbol{x}_{\text{target}, i})),
\end{align}
where $r(a,b)$ denotes the Pearson correlation coefficient between $a$ and $b$, and $abs(\cdot)$ denotes the absolute coefficient. The absolute value is adopted to focus on the magnitude of correlations. Essentially, the larger the two coefficients are, the easier the adversary can learn the feature correlations via GRNA.

Fig.~\ref{fig-corre} shows the correlations computed by Eqn~(\ref{eq-adv}) and~(\ref{eq-yhat}). The fractions of $d_{\text{target}}$ used for the bank marketing (in Fig.~\ref{fig-bank-corr}) and credit card (in Fig.~\ref{fig-credit-corr}) datasets are 40\% and 30\%, respectively. In addition, the x-axis represents the MSE for each feature in $\boldsymbol{x}_{\text{target}}$.
We can observe that the correlations \textit{w.r.t.} both $\boldsymbol{x}_{\text{adv}}$ and $\boldsymbol{v}$ impact the attack performance of GRNA.
A weaker correlation between $\boldsymbol{x}_{\text{target}, i}$ with $\boldsymbol{x}_{\text{adv}}$ and $\boldsymbol{v}$ results in a lower inference accuracy, such as features 1 and 3 in Fig.~\ref{fig-bank-corr} and features 4 and 6 in Fig.~\ref{fig-credit-corr}.
The rationale is that a weak correlation implies that the change of the unknown feature value only has a minor impact on the generated sample and the prediction output; thus, its value range is broader than the other unknown features, leading to an inaccurate inference.
%


Another insight is that GRNA can achieve different reconstruction accuracy regarding different features. The MSE metric computed by Eqn~(\ref{eq:mse-per-feature}), which averages the reconstruction errors on all target features, is mainly used to evaluate the overall performance of our attacks. In real-world applications, we should note that features closely correlated to $\boldsymbol{x}_{\text{adv}}$ can be more precisely reconstructed than the relatively independent ones. For example, after obtaining the inferred deposit and shopping information based on a user's income, the adversary may be more convinced of the inferred deposit as it is more relevant to the income feature.
\section{Countermeasures}\label{sec:defense}

In this section, we discuss several potential defense methods that may mitigate the proposed feature inference attacks.

\vspace{1mm}\noindent
\textbf{Rounding confidence scores.}
In ESA, the adversary relies on the exact linear equations for the inference. Thus, a possible defense to ESA is to coarsen the confidence scores $\boldsymbol{v}$ returned to the active party, for example, round $\boldsymbol{v}$ down to $b$ floating point digits before revealing it.
Since the performances of the two random guess methods are similar, we only include the random guess with uniform distribution in this set of experiments.

Fig.~\ref{subfig:bank-conf-esa}-\ref{subfig:drive-conf-esa} show the effect of this strategy on two datasets, respectively.
We can observe that when rounding to 0.1 (i.e., $b=1$), the MSE per feature is higher than the random guess method, and the result is relatively stochastic. This is because that the equation result in Eqn~(\ref{eq:single-lr-equation}) or~(\ref{eq:converted-equations}) is related to $\ln \boldsymbol{v}$, a change of $\boldsymbol{v}$ with respect to 0.1 has thus a big impact on that result, leading to an inaccurate inference. In contrast, rounding to 0.001 (i.e., $b=3$) only has a small impact because the result would be mainly determined by the former three floating point digits.
Fig.~\ref{subfig:bank-conf-grn}-\ref{subfig:drive-conf-grn} show the effect of the rounding strategy in GRNA for the LR model.
The results illustrate that GRNA is insensitive to the rounding of confidence scores, and the adversary can obtain a similar performance comparing to that without rounding. The reason is that GRNA learns the overall correlation between the adversary's features and the attack target's features, and the low-precision prediction outputs still indicate this pattern.

\vspace{1mm}\noindent
\textbf{Dropout for neural networks model.} Overfitting is considered to be an important factor in several inference attacks in the training stage \cite{ShokriSSS17, NasrSH2019}, as the trained model may memorize some underlying information of the data. We utilize a state-of-the-art regularization technique called dropout \cite{SrivastavaHKSS14} for training the vertical NN model to avoid overfitting. As depicted in Fig.~\ref{subfig:credit-dropout-grn}-\ref{subfig:news-dropout-grn}, using dropout slightly leads to a higher MSE per feature (i.e., degrade the attack performance), because dropout encourages NN models to memorize less distribution information of underlying datasets. Nevertheless, the adversary can still have a good inference because the generator model learns the overall feature correlations.

\vspace{1mm}\noindent
\textbf{Pre-processing before collaboration.}
Based on the experimental results in Section~\ref{sec:experiments}, we observe that:
(i) if the number of classes is relatively large, ESA can precisely recover the attack target's feature values, and PRA can correctly infer most branches in the tree model; (ii) if the adversary's features are highly correlated with the attack target's features, the attack performance of GRNA could be greatly improved.
These observations inspire the parties to execute a pre-processing step that mitigates the potential privacy risks before data collaboration.
First, the parties check the relationship between the number of classes and the number of their contributed features, ensuring no obvious privacy vulnerabilities.
Second, the parties collaboratively execute a secure protocol (e.g., using secure multiparty computation \cite{WuCXCO20}) to compute their feature correlations and remove those closely related features.

\vspace{1mm}\noindent
\textbf{Post-processing for verification.} The parties can also execute an additional verification step to check if the prediction output would lead to privacy leakage before revealing to the active party.
For example, if the parties utilize secure hardware, e.g., Intel SGX, for computing the model predictions, then they can mimic these attacks inside the secure enclaves. Specifically, if the possible leakage exceeds a pre-defined threshold for any party, they do not reveal the prediction output for privacy protection.
In other words, the prediction output is released only when it satisfies the privacy of all parties.
However, this post-processing step may incur huge overheads to the computation of model predictions.

\vspace{1mm}\noindent
\textbf{Hide the vertical FL model.}
Another consideration is to hide the trained vertical FL model using ciphertexts or secure multiparty computation compatible values, such that the adversary does not have access to the plaintext model in the attack.
This method could mitigate these attacks, but the active party cannot justify if the trained model is reasonable or not and cannot have a good interpretation of the prediction output for making important decisions.
A possible alternative is to use explainable predictions instead of revealing the trained vertical FL model, such that the contribution of each feature is provided to the active party for justification. However, it is still an open problem whether those explainable predictions would cause new privacy leakages.



\vspace{2mm}\noindent
\textbf{Differential Privacy (DP).} DP~\cite{XiaoWG10} is a state-of-the-art privacy protection tool. However, it is unsuitable in our setting.
In particular, DP requires that any information from sensitive data should be released using a randomized function, such that even if we arbitrarily change a record in the function's input, the distribution of the function's output remains roughly the same.
In our context, the sensitive data is an unlabelled record, and the function that releases information is a machine learning model that predicts the record's label.
If we are to achieve DP, then we should inject noise into the model, such that even if we arbitrarily modifies the input record, the distribution of the model's output label is almost unchanged.
In other words, the model should provide roughly the same prediction for any input record.
This apparently renders the model useless. Therefore, DP is unsuitable for our problem.


\section{Related Work}\label{sec:related-work}

\vspace{1mm}\noindent
\textbf{Vertical federated learning.} 
With horizontal FL being thoroughly studied~\cite{McMahanMRHA17, choudhury2020anonymizing, YangLCT19}, vertical FL is receiving increased attention, and there have been several solutions~\cite{hardy2017private, Mohassel17, OhrimenkoSFMNVC16, ChengCorr19, HuNYZ19, WuCXCO20} proposed recently. 
\cite{hardy2017private, ChengCorr19} apply partially homomorphic encryption \cite{Paillier99} in the LR and GBDT models to protect the information exchanged among the parties.
\cite{Mohassel17} and \cite{WuCXCO20} utilize secure multiparty computation to provide strong data security for the NN and tree-based models, respectively, ensuring that no intermediate information other than the final output is disclosed.
%
Also, secure hardware could protect the parties' private data during the training and prediction stages using secure enclaves \cite{OhrimenkoSFMNVC16}. 
By allowing that the active party's labels can be shared with the passive parties, \cite{HuNYZ19} proposes a vertical FL framework, by which each party computes a prediction from a trained local model, and then the parties aggregate the local predictions into a final prediction. 

Nevertheless, existing vertical FL solutions focus on protecting data privacy during the training or prediction process. In this paper, we consider the most stringent case that the computation of model training and model prediction is secure enough, and the adversary only utilizes the computation output (i.e., the trained model and model predictions) to infer the attack target's private feature values. 

\vspace{1mm}\noindent
\textbf{Inference attacks on federated learning.}
Recent studies have demonstrated that FL is vulnerable to multiple types of inference attacks, such as \textit{membership inference}~\cite{NasrSH2019, MelisSCS19}, \textit{property inference}~\cite{MelisSCS19}, and \textit{feature inference}~\cite{hitaj2017deep, ZhuLH19, wang2019beyond}. 
Membership inference~\cite{NasrSH2019, MelisSCS19} aims to determine whether a specific record is in a party's training dataset or not, and the attacking effectiveness is closely related to the overfitting nature of ML algorithms~\cite{ShokriSSS17, Salem0HBF019}. 
However, this attack does not exist in vertical FL because every party knows all the training sample ids intrinsically. 
Property inference~\cite{MelisSCS19} attempts to extract some underlying properties or statistics of a party's training dataset, which are uncorrelated to the training task~\cite{ganju2018property, MelisSCS19}. 
Feature inference~\cite{hitaj2017deep, ZhuLH19, wang2019beyond} is to recover the samples used in a party's training dataset. 

Unfortunately, these inference attacks are only applicable to the training stage of horizontal FL. The reason is that they greatly rely on the iteratively exchanged gradients during the training process to build meta-classifiers for the inference. 
On the contrary, we focus on the feature inference in the prediction stage of vertical FL, which is more challenging because the adversary knows little information about the attack target's data. To our knowledge, this is the first work that addresses this problem. In addition, our attack methods do not depend on any intermediate information during the prediction process. 



\vspace{1mm}\noindent
\textbf{Other related studies.}
In addition to feature inference in FL, another line of research tries to infer private feature values based on model predictions of a centralized ML model. For example, \cite{fredrikson2014privacy} proposes an attack to infer patients' genetic markers given black-box access to a linear regression model. Nevertheless, the adversary needs background information of all features except the targeting feature, making this attack relatively impractical. Similarly, the inference attack to the DT model in~\cite{fredrikson2015model} requires a marginal prior distribution for each feature, and the image reconstruction attack also relies on auxiliary information to define the cost function of the attack.
In vertical FL, it is often difficult for the active party to collect prior information from the passive parties. In contrast, our attack methods do not rely on any background information of the attack target's data distribution.

The basic idea of~\cite{zhang2020secret} is similar in spirit to ours, i.e., inferring sensitive features (pixels) based on known image patches and white-box models. But this scheme requires pre-training on public images for obtaining the general image distributions, whereas our approach does not rely on the prior data distribution of the target features. Besides, \cite{zhang2020secret} aims to recover the training samples, whereas our attack methods mainly focus on reconstructing the prediction samples.
\section{Conclusion}\label{sec:conclusion}

We present two specific attacks based on individual prediction output: equality solving and path restriction, for logistic regression and decision tree models, respectively. Furthermore, we design a general attack based on multiple prediction outputs for neural networks and random forest models. To our knowledge, this is the first work that investigates the privacy leakages in vertical FL. The experimental results demonstrate the accuracy of the proposed attacks and highlight the need for designing private algorithms to protect the prediction outputs. 
\balance
\bibliographystyle{IEEEtran}
\bibliography{sigproc}

\vspace{12pt}

\end{document}